\documentclass{article}
\newcommand{\etal}[0]{\textit{et al.}}

\usepackage{arxiv}

\usepackage[utf8]{inputenc} 
\usepackage[T1]{fontenc}    
\usepackage{hyperref}       
\usepackage{url}            
\usepackage{booktabs}       
\usepackage{amsfonts}       
\usepackage{nicefrac}       
\usepackage{microtype}      
\usepackage{lipsum}
\usepackage{graphicx}

\graphicspath{ {./images/} }

\title{Responsible Personalisation: The Double-Edged Sword of Personalisation in Human-Robot Interaction}

\author{
  Antonio Andriella\thanks{These authors contributed equally to this work, where the names are written in alphabetical order of the last name} \\
  Institut de Robòtica i Informàtica Industrial (CSIC-UPC) \\
  Llorens i Artigas 4-6, 08028, Barcelona, Spain \\
  \texttt{aandriella@iri.upc.edu} \\
  \And
  Jauwairia Nasir\footnotemark[1] \\
  Universität Augsburg \\
  Augsburg, Germany \\
  \texttt{jauwairia.nasir@uni-a.de} \\
  \And
  Andrea Rezzani \\
  Free University of Bozen-Bolzano \\
  Bozen, Italy \\
  \texttt{anrezzani@unibz.it} \\
  \And
  Alyssa Kubota \\
  San Francisco State University \\
  San Francisco, CA, USA \\
  \texttt{akubota@sfsu.edu} \\
  \And
  Dimitri Lacroix \\
  Nantes Université, Univ Angers, Laboratoire de psychologie des Pays de la Loire, LPPL \\
  UR 4638, F-44000, Nantes, France \\
  Bielefeld University, \\
  Center for Cognitive Interaction Technology (CITEC)
  \\ Germany \\
 \texttt{Dimitri.lacroix@univ-nantes.fr} \\
  \And
  Tamlin Love \\
  Institut de Robòtica i Informàtica Industrial (CSIC-UPC) \\
  Llorens i Artigas 4-6, 08028, Barcelona, Spain \\
  \texttt{tlove@iri.upc.edu} \\
  \And
  Aniol Civit \\
  Institut de Robòtica i Informàtica Industrial (CSIC-UPC) \\
  Llorens i Artigas 4-6, 08028, Barcelona, Spain \\
  \texttt{acivit@iri.upc.edu} \\
  \And
  Vicky Charisi \\
  Singapore-MIT Alliance for Research and Technology \\
  Singapore \\
  Centre for Collective Intelligence, MIT \\
  Cambridge, MA, USA \\
  \texttt{vasiliki.charisi@smart.mit.edu} \\
  \And
  Elisabeth Andre \\
  Universität Augsburg \\
  Augsburg, Germany \\
  \texttt{elisabeth.andre@uni-a.de} \\
  \And
  Wing--Yue Geoffrey Louie \\
  Oakland University \\
  Rochester, MI, USA \\
  \texttt{louie@oakland.edu}
}

\begin{document}
\maketitle

\newpage

\begin{abstract}
While personalisation is becoming a defining capability in human–robot interaction (HRI), the existing literature on responsible personalisation remains fragmented, offering isolated accounts of ethical risks without a structured understanding of how they emerge across interaction contexts. This gap is particularly critical in HRI, where robots' embodiment and social presence can amplify and reshape such risks or generate new types of risks.
We present a lifecycle-based and context-sensitive framework for personalised HRI, grounded in an embodiment-aware perspective. The framework combines stages of the personalisation process with interaction characteristics (short- vs. long-term, open- vs. closed-domain), enabling systematic analysis of how risks arise and evolve.
Building on this, we conduct an integrative analysis of key ethical risks, including autonomy erosion, biased user modelling, manipulation, dehumanisation, and privacy violations, and examine how they manifest across contexts. We translate these insights into actionable design recommendations and outline open research challenges.
By structuring both the design space and risk landscape of personalised HRI, this work provides a foundation for more systematic, transparent, and ethically grounded approaches to personalised robot behaviour.
\end{abstract}



\section{Introduction}
\label{sec:introduction}

Personalisation has become a cornerstone in Human-Robot Interaction (HRI), enabling robots to adapt their behaviour to individual users in ways that foster engagement, task effectiveness, as well as trust and collaboration among the parties \cite{ahmad_systematic_2017, Rossi_PR17, yang_impact_2024}. Across domains such as healthcare, education, and hospitality, personalised interactions have shown clear benefits, both in short-term encounters, where quick adaptation is key~\cite{Andriella_IJSR19}, and in long-term relationships, where the responsiveness to users' evolving needs is crucial~\cite{Irfan_hri26}.

\begin{figure*}[tbp]
    \centering
    \includegraphics[width=\linewidth]{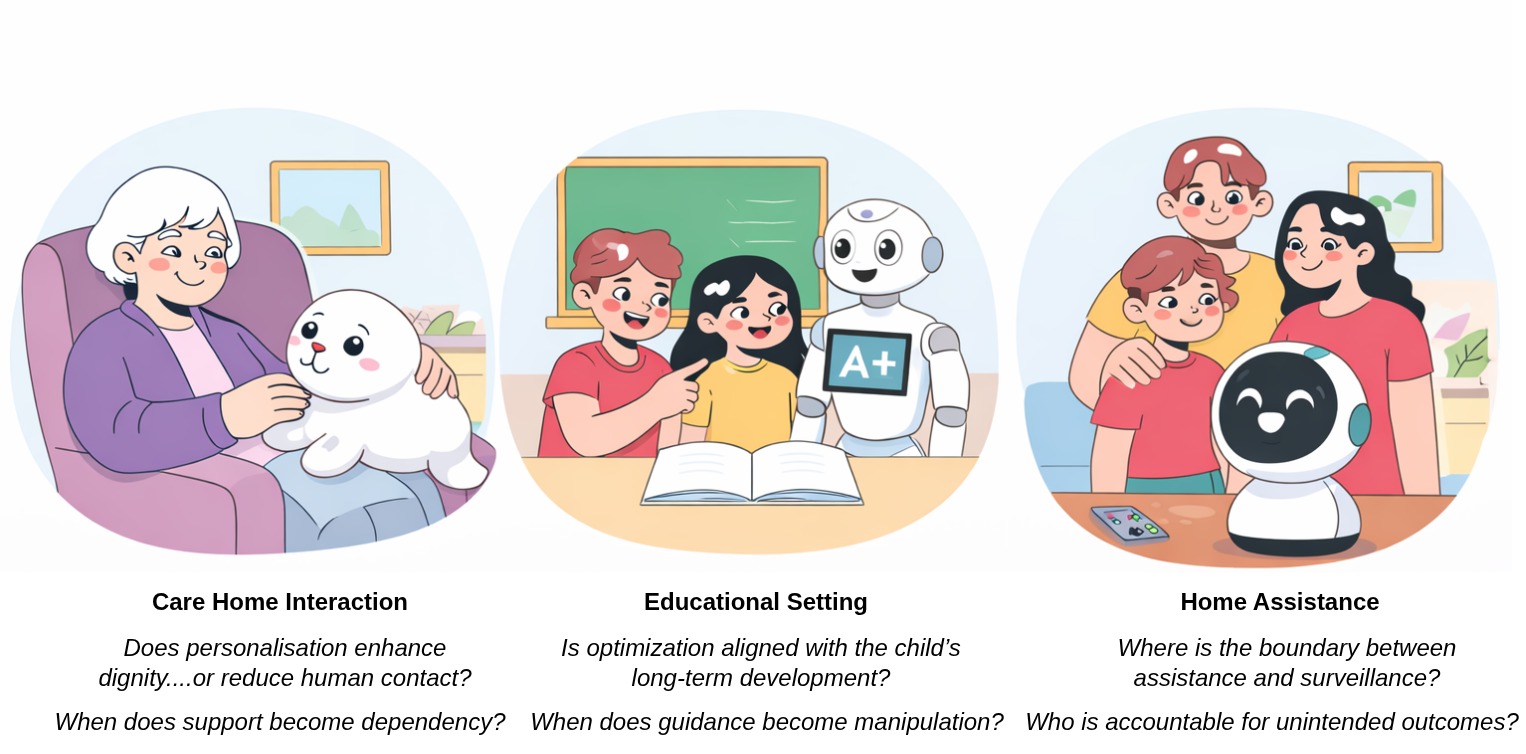}
    \caption{Relevant questions in common scenarios in HRI that require critical reflection when designing for personalisation.} \label{critical_questions}
\end{figure*}

That said, there are reasons to believe that personalisation in HRI is a double-edged sword. While it promises to enhance user experiences  \cite{yang_impact_2024, lacroix_making_2025}, it also poses significant risks such as potentially undermining user autonomy and amplifying societal biases (see Figure \ref{critical_questions}). These tensions have been increasingly acknowledged within the Human-Computer Interaction (HCI) community, where critical research has addressed issues of loss of agency, manipulation, stereotyping, and ethical design \cite{treiblmaier_evaluating_2004, berkovsky_influencing_2012, sundar_personalization_2010, limerick_experience_2014, mittelstadt_auditing_2016, Mittelstadt2016}; however, these concepts have not yet been examined systematically or remain under-explored within the relatively new HRI community. While some risks are transferable between the two domains, the challenges that come due to the embodied nature of the robots in HRI may introduce unique or amplified risks that are distinct from the discourse in HCI. This can include challenges with maintaining privacy~\cite{Stapels_IJSR23} as a robot is mobile while being co-located in the same space as a human, increased risks of manipulation~\cite{Bertolini_pt22}, as robots have, in some studies, been found to be more persuasive than computers \cite{Deng2019}, and increased risk of physical harm as they can manipulate and act on the physical environment \cite{Perlo_2025}. 

With the aim of further exploring this research gap, we organised a first workshop at the \textit{removedfordoubleblind} conference in 2023\footnote{\url{https://sites.google.com/view/warn-roman23/home}}. The outcomes of this initial event, combined with a review of the current state of the art, highlighted both the community's interest in the topic and the lack of systematic reflection on the risks associated with personalisation in HRI. From this, we identified several key themes as especially pressing for the community: (1) the potential erosion of user autonomy, (2) the risk of reinforcing societal stereotypes, and (3) privacy violations. These themes became the focal point of a second and a third workshop at the \textit{IEEE International Conference on Robot and Human Interactive Communication} conferences (RO-MAN) in 2024\footnote{\url{https://sites.google.com/view/bailar2024/home}} and 2025\footnote{\url{https://bear-workshop.github.io/website/}}, respectively, structured around four Oxford-style debates, each focused on one of the aforementioned themes. The debates brought together sixteen experts from diverse fields such as robotics, psychology, sociology, philosophy, and law. The discussions during the debate and those that followed afterwards revealed a set of challenges that, while highly relevant, remain underexplored in HRI. This paper draws on these discussions to articulate the risks of personalisation with social robots and to offer design-oriented recommendations to support a more responsible and mindful approach to personalisation. Note that although this work also draws conceptual inspiration from HCI research on responsible personalisation, it does not seek to systematically review the broader HCI literature on the topic.

More concretely, in this paper: In Section \ref{sec:foundations}, we first identify the \textbf{embodiment-driven affordances} of robots that could amplify or introduce unique challenges for personalisation in HRI. Next, we briefly present how personalisation is defined in HRI and its \textbf{typical lifecycle} to set the stage for the discourse; followed by a \textbf{``loose classification framework''} for social HRI, differentiating between short- vs long-term interactions and open- vs closed-domain scenarios, to highlight how context could shape the manifestation of risks. Then, in Section \ref{sec:risks_of_personalisation}, we analyse the \textbf{main risks of personalisation}, and we contextualise them within the lifecycle and classification proposed in Section \ref{sec:foundations}. Lastly, in Section \ref{sec:recommendations}, we present a set of \textbf{design recommendations} for the identified risks to serve as practical guidelines for the concerned stakeholders, as well as propose open research questions as an invitation to the community to work towards understanding and designing truly responsible personalised robots.

\begin{figure*}[tbp]
    \centering
    \includegraphics[width=0.7\linewidth]{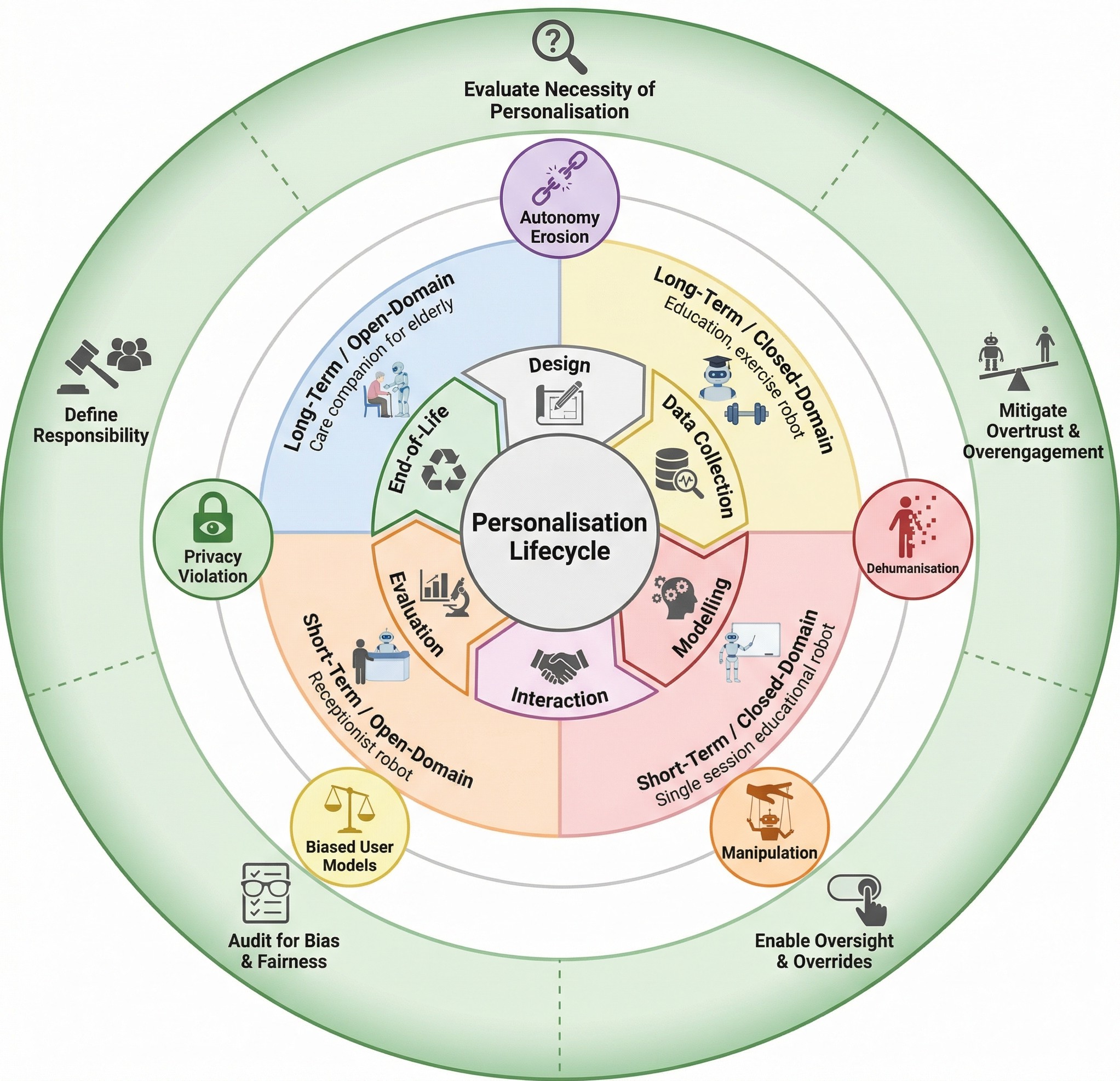}
    \caption{Overview of the responsible personalisation framework that situates personalisation within a lifecycle perspective and highlights how contextual factors, ethical risks, and mitigation strategies interact in HRI systems.} \label{Summative_figure}
    \label{system_view_RP}
\end{figure*}

All in all, the primary contribution of this paper is the \textbf{Responsible Personalisation Framework} (see Figure \ref{system_view_RP}), which comprises three interrelated contributions that vary in scope and depth (written here in descending order of impact):
 
\begin{itemize}
    \item \textbf{An integrative risk analysis linked to actionable design guidance:} Building on interdisciplinary insights and workshop discussions, we identify key ethical risks, such as autonomy erosion, biased user modelling, manipulation, dehumanisation, and privacy violations, and connect them to concrete, design-oriented recommendations and open research questions for responsible development.
    \item \textbf{A life-cycle based and context-sensitive analytical framework for personalisation:} We introduce a unified perspective that combines a personalisation lifecycle with a classification of interaction types (short- vs. long-term, open- vs. closed-domain), enabling systematic analysis of how risks emerge, evolve, and persist across different HRI settings.
    \item \textbf{Conceptualisation of responsible personalisation in HRI through an embodiment-aware lens:} We articulate how personalisation in HRI could differ from HCI by foregrounding the role of embodiment, showing how physical presence has the potential to amplify both the benefits and ethical risks of adaptive behaviour.
\end{itemize}

Bringing the core concepts introduced in this paper, we provide a system view of \textit{Responsible Personalisation} visually in Figure \ref{system_view_RP}. The framework conceptualises personalisation as a lifecycle comprising six interconnected phases: design, data collection, modelling, interaction, evaluation, and end-of-life considerations. This lifecycle is applicable across diverse HRI contexts; however, the manifestation and implications of each phase may vary according to contextual factors such as interaction duration (short-term versus long-term) and application domain (open versus closed) that shape both the opportunities afforded by personalisation and the risks that emerge throughout the personalisation lifecycle in HRI systems. In parallel, the framework identifies key ethical risks, including privacy violations, autonomy erosion, manipulation, dehumanisation, and biased user models, and maps them to responsibility-oriented mitigation strategies. Collectively, the framework provides a structured foundation for designing and evaluating personalised robotic systems that are both effective and ethically aligned.

Recent work by Axelsson and Seeck \cite{Axelsson_and_Seeck_2026} on social robotic surveillance and manipulation similarly highlights risks associated with pervasive data collection, personalisation, and robotic influence, particularly through the lenses of surveillance capitalism and governmentality. However, while their contribution primarily provides a socio-political and critical-theoretical analysis of surveillance and manipulation in social robotics, our work focuses on the design and operational dimensions of responsible personalisation in HRI. Specifically, we contribute a lifecycle-based and context-sensitive analytical framework of responsible personalisation that systematically maps how personalisation-related risks emerge across different stages of system development and interaction contexts, while also providing actionable design-oriented recommendations for responsible personalised HRI. Taken together, this work is positioned \textit{responsible personalisation} as a critical and timely challenge for the HRI community. By bridging conceptual clarification, risk analysis, and practical guidance, the paper provides a structured starting point for researchers and practitioners seeking to design, evaluate, and deploy personalised robotic systems more responsibly. Rather than advocating for or against personalisation, we aim to enable a more reflective and context-sensitive approach - one that recognises both its potential and its ethical complexity - and that supports the development of human-centred, accountable, and responsible personalised robotic systems.

\section{Foundations for Responsible Personalisation in HRI}
\label{sec:foundations}
To systematically analyse the risks and ethical implications of personalisation in HRI, it is first necessary to establish a shared conceptual foundation. Personalisation in HRI does not emerge from a single mechanism, but rather from the interplay between the robot's embodied nature, the ways personalisation is defined and operationalised, the lifecycle processes through which adaptive behaviour unfolds, and the social interaction contexts in which robots are deployed. Together, these elements shape how personalised behaviour is perceived, enacted, and experienced over time.

This section, therefore, introduces the foundational concepts that underpin our Responsible Personalisation Framework. First, we discuss the embodiment-driven affordances of robots and explain how physical embodiment may amplify or introduce unique opportunities and risks compared to disembodied AI systems (see Section~\ref{sec:embodimentsection}). Next, we clarify the terminology surrounding customisation, adaptation, and personalisation in HRI, establishing a shared vocabulary for the remainder of the paper (see Section~\ref{sec:definitions}). Following this, we present a lifecycle-based perspective that conceptualises personalisation as an ongoing process spanning design, data collection, modelling, interaction, evaluation, and end-of-life considerations (see Section~\ref{sec:mechanics_of_personalisation}). Finally, we introduce a classification of social HRI contexts based on interaction duration and domain specificity, highlighting how different interaction settings shape the manifestation and persistence of personalisation-related risks (see Section~\ref{sec:types_social_hri}).


    \subsection{Embodiment Driven Affordances of Robots}
    \label{sec:embodimentsection}
    In HRI, physical embodiment refers to the robot's existence as a physically instantiated agent situated within the user's environment and capable of acting through a material body~\cite{Wainer_roman06, bartneck_human-robot_2020}. Unlike disembodied AI systems, such as conversational agents or screen-based interfaces, embodied robots occupy physical space, express behaviour through movement, gaze, posture, gestures, and spatial positioning, and can directly manipulate or affect the surrounding environment. Importantly, embodiment is not merely a physical property, but also a social and perceptual one: users often interpret embodied robots as intentional and socially present agents, attributing agency, emotions, competence, or authority to them during interaction~\cite{Navare_SR24, Spatola_CHB21, thellman_mental_2022}.

    
    In this section, we present evidence suggesting that embodiment can amplify or introduce unique challenges for personalisation in HRI, stemming from differences in how users perceive and respond to embodied agents. To begin with, when it comes to interaction experience, a systematic review by Deng~\etal\cite{Deng2019} consisting of 65 studies reports that physical embodiment enhances users' perceptions of an agent relative to virtual counterparts and can also improve task performance, particularly when the robot assumes either a superior or subordinate role. Further, Kwak~\etal\cite{Kwak2013} found that children were significantly more empathetic toward a physically embodied robot than toward its on-screen virtual counterpart. Similarly, Lee~\etal\cite{LEE2006} reported that people evaluated the robot, their interactions with it, and its social presence more positively when it was physically embodied rather than presented as a screen-based character. 
    
    When it comes to trust, several studies suggest a nuanced picture. Some have found no significant differences in trust ratings between a robot and its virtual counterpart \cite{Maris2017} or a smart speaker \cite{Dimosthenis2020}. However, in a former longer-term study, Van Maris~\etal\cite{Maris2017}, participants tended to trust the robot more in later sessions than in earlier ones, indicating a significant temporal effect in the development of trust toward an embodied agent and raising the possibility of embodied persuasion. Similarly, the latter study  by Kontogiorgos~\etal\cite{Dimosthenis2020} reports that participants were more willing to continue using a robot than a speaker following system failures. 
    On the other hand, in healthcare contexts, participants have indeed clearly been shown to follow instructions, place greater trust in, and engage more with physically embodied robots compared to tablet-based interfaces \cite{Jordan2015}. Similarly, in the context of game companionship \cite{Kim2025}, interactions with physically embodied robots elicited higher arousal and a stronger preference for future engagement than interactions with depicted robots (wall projection of the robot). However, within the same domain and focusing on children, prior work found no significant effects of embodiment on subjective measures such as attraction, involvement, or perceived appearance; the only notable difference was that physically embodied agents were rated as more enjoyable when gestures were included \cite{KoseBagci2009}. 
    
    Taken together, these findings illustrate that physical embodiment actively amplifies the social, cognitive, and affective effects of artificial agents. This amplification effect becomes particularly consequential when personalisation is introduced. Leyzberg~\etal\cite{Leyzberg2012} showed that a physically present, personalised robot tutor led to greater cognitive learning gains than non-physical (voice-only) or virtual personalised tutoring formats. The same adaptive behaviours that may be perceived as helpful or engaging in disembodied systems can acquire heightened persuasive power, social presence, authority, and intentionality when enacted through an embodied robot. 
    
    As a result, personalisation in HRI is uniquely shaped by users' interpretations of a robot's physical presence, social cues, and perceived agency - interpretations that vary widely across individuals, contexts, and over time. Embodiment thus interacts with personalisation along at least two dimensions. First, it might increase salience: personalised adaptations can be more noticeable in both physical and social contexts, making them potentially more difficult to overlook or disengage from. Second, it might strengthen agency attribution: users could be more inclined to interpret personalised behaviour as intentional or caring, which may foster trust but also lead to greater susceptibility to influence. Although these embodiment-driven affordances do make personalisation more effective, at the same time, they can magnify harm. As such, embodiment complicates assumptions about when, how, and for whom personalisation is desirable in HRI, underscoring the need for greater conceptual clarity about what personalisation entails in HRI and how it is operationalised across systems. In the next section, we address this need by clarifying foundational terminology and introducing a lifecycle-based framework for personalisation in HRI.

    \subsection{Definitions} \label{sec:definitions}

    To establish a clearer conceptual foundation, we distinguish the terms \textit{customisation}, \textit{adaptation}, and \textit{personalisation} based on three key dimensions: (i) who initiates the change of robot behaviour, (ii) how the change is implemented, and (iii) the scope and dynamics of that change.
    These distinctions are particularly important in HRI because the distribution of control between user and robot directly affects issues such as autonomy, transparency, and the potential for manipulation. The following definitions clarify how each concept operates within the broader landscape of personalised robot behaviour.

    \subsubsection{Customisation}
    
    Customisation refers to modifications of a robot's behaviour, appearance, or interaction parameters that are explicitly specified by the user~\cite{orji_comparison_2017, lacroix_whos_2023, sundar_personalization_2010}. 
    Customisation typically occurs at the beginning of the interaction or through dedicated configuration interfaces. Examples include selecting the robot's voice, choosing preferred interaction modalities (e.g., text vs. speech), or setting notification frequencies~\cite{Leonardi_hci19}. These changes are therefore intentional and potentially more transparent, as they result directly from user decisions rather than from system inference.
    Customisation puts the responsibility of system configuration on stakeholders, including both primary users and other involved parties, to configure the robot according to their preferences. This can hinder robot adoption as it can increase cognitive effort and training times, and disadvantage users who lack the technical expertise, time, or confidence to adjust system settings effectively.    
    For this reason, customisation is often used in combination with other forms of adaptation in HRI systems to complement it by adjusting interaction parameters dynamically when explicit configuration is impractical or insufficient.

    In brief, customisation is mainly \emph{user-initiated}, \emph{executed through direct user input} such as explicit configuration choices, and \emph{mostly scoped to a fixed set of tunable parameters} whose values change only when the user revisits them. 
    
    \subsubsection{Adaptation}
    Adaptation refers broadly to a robot's capacity to modify its behaviour in response to changes in its environment, interaction context, or internal system state~\cite{Rossi_PR17}. In HRI, adaptation encompasses personalised responses, for example, a robot lowering its voice in a noisy room (contextual adaptation)~\cite{Tuttosi_iros23} or turning to face a speaker (social adaptation)~\cite{Skantze_hri25}.
    Importantly, adaptation does not necessarily imply modelling a specific individual. In many cases, adaptive behaviour can be stateless, relying solely on immediate contextual cues. Such mechanisms enable robots to react to situational changes without maintaining a persistent representation of the user. They are often implemented through rule-based policies~\cite{Sanders_sai16}, reactive control strategies~\cite{Foster_IJSR17, Andriella_IJSR19}, or models trained on population-level data that approximate behaviour suitable for a general class of users~\cite{Delduchetto_RAL22, Nasir2024}. 
    Because these systems are designed to operate for a broad population, they often reflect assumptions about an ``average'' user. While this approach can support robustness and usability in diverse environments, it may also obscure individual differences and reproduce biases embedded in training data or design assumptions~\cite{Huang_hri2025}. Consequently, adaptive behaviour can improve interaction fluency while still failing to account for the needs, preferences, or abilities of specific individuals.

    A way to address this issue is by incorporating varying degrees of human involvement, giving rise to human-in-the-loop adaptation, which could make it more suited to the users. This form of adaptation allows the human to guide or validate the robot's learning process at key points through explicit feedback. This approach can help preserve user autonomy, improve transparency, and reduce the risk of bias or over-reliance, while still benefiting from adaptive system capabilities. Examples include co-adaptive learning and shared control~\cite{Baraka_THRI25}.
    However, this approach is not without drawbacks. The requirement for human participation at key decision points of the adaptation process may limit scalability and slow down system responsiveness. Additionally, human feedback can be contradictory and may evolve over time, leading to inconsistencies. In addition, it remains unclear whether the user's engagement decreases as time passes or if their ongoing involvement is continuously necessary.
    Finally, the integration of human input introduces additional data privacy and accountability challenges, as decisions are shaped by both the system and identifiable individuals.

    In brief, adaptation in its general form is \emph{robot-initiated}, \emph{implemented through the system's response to contextual or population-level cues rather than a model of the individual}, and \emph{dynamic in the moment} but without necessarily accumulating a persistent representation of a specific user.

    \subsubsection{Personalisation}
    In contrast to general user-driven customisation and adaptation, personalisation refers to the process by which a robot system autonomously learns about and adjusts its behaviour based on information inferred about a specific individual user~\cite{Rossi_PR17, lacroix_whos_2023, sundar_personalization_2010}. 
    While adaptation typically responds to contextual cues or population-level modelling without modelling a particular individual, and customisation relies on explicit user configuration, personalisation extends adaptation by incorporating a representation of the specific user into the adaptation process.
    This representation, commonly referred to as a \emph{user model}, captures characteristics such as preferences, behavioural tendencies, abilities, or needs. By maintaining and updating this model over time, the robot can tailor its behaviour to a specific individual rather than to a generic or population-level user profile. In this sense, personalisation enables a more targeted form of adaptation that accounts for stable and evolving characteristics of the user.

    Personalisation is typically system-initiated, often driven by machine learning or AI algorithms, and may involve both implicit and explicit data collection~\cite{Andriella_Springer25}. Personalisation can occur in real-time, through online learning and immediate behavioural adaptation~\cite{Garcia_RAL21}, or offline, where models are trained from previously collected data or updated between interactions before being redeployed~\cite{Park_aaai19}. This process may target real-time changes in the robot's behaviour, updates to the underlying user model, or both.
    In HRI, personalised behaviour can manifest across multiple levels of interaction. A robot tutor may adapt its feedback style based on a child's learning pace~\cite{Leyzberg_THRI18}, while an assistive robot may adjust its reminders to align with an older adult's daily routine~\cite{DiNapoli_UMUAI23}. Importantly, personalisation tends to operate dynamically, often without the user's direct (explicit) input. In this sense, a significant distinction from adaptive systems is the shift of control toward the relationship between the system and the user, where the system seeks to enhance its understanding of the user in order to best align its behaviour. This creates ethical implications related to transparency~\cite{Kraus_umap22}, autonomy~\cite{Collier_hri25}, and data privacy~\cite{Khaksar_TFSC24}.

    In brief, personalisation can be defined as \emph{system-initiated}, but, in contrast with adaptation, is \emph{implemented through the robot inferring and learning about a specific individual}, and is \emph{scoped to a persistent, evolving user model} that is updated across the interaction and tailored to that individual.

    \noindent \textbf{Summary.} 
    Taken together, these distinctions are best understood as regions along a continuum of control and inference rather than as discrete categories. Real systems frequently combine them: a robot may be configured by the user, adapt to its context, and personalise to an individual within a single interaction, and human-in-the-loop mechanisms cut across the boundary entirely. We therefore do not propose these terms as a strict taxonomy, but as a shared vocabulary for reasoning about where control resides and how behavioural change is produced.

    \subsection{The Lifecycle Mechanics of Personalisation}
    \label{sec:mechanics_of_personalisation}
    
    To fully understand how personalisation operates in HRI, it is useful to integrate both temporal and functional perspectives. We therefore frame personalisation as a six-phase lifecycle (temporal), while concurrently highlighting its underlying input–modelling–output (IMO) mechanics (functional), where: (i) the \textit{input layer} outlines what the robot observes and learns from, (ii) the \textit{modelling layer} defines what the robot infers based on that input, and (iii) the \textit{output layer} outlines what the robot changes about its behaviour. This dual framing helps us trace how the robot learns, interprets, and adapts over time. Although described separately for clarity, lifecycle phases and IMO components continuously interact throughout the development and deployment of personalised HRI systems.
    
        \subsubsection{Design - Establishing the foundations of personalisation}
        The lifecycle begins during the \emph{Design} phase, where developers define the purpose, scope, and boundaries of personalisation. Here, the future input layer is specified (what the robot may observe), the intended modelling layer is planned (which user characteristics and their relationships may be inferred), and the potential output layer is constrained (which behaviours may adapt)~\cite{Rossi_PR17}.  
        In HRI, design choices include tasks, deployment settings, sensing channels, robot morphology, social behaviours, roles, degrees of initiative, and ethical principles~\cite{Onnasch_IJSR21, bartneck_human-robot_2020} related to inference, autonomy, and transparency~\cite{Barajas_Springer25, Pollmann_MDPI21}. These early decisions constrain the forms of personalisation that are feasible and also which risks may arise long before deployment~\cite{Pollmann_TFSC23, vanMaris_FRAI2020, Callari_TS24}.
        
        Because they fix the system's degree of initiative, its sensing and inference boundaries, and the objectives it optimises for, these design decisions are not merely enabling conditions but the point at which several of the risks, from autonomy erosion to manipulation and privacy exposure, first take root, long before any interaction occurs.
        
        \subsubsection{Data Collection - What the robot learns from}
        During the \textit{Data Collection} phase, the \textit{input layer} is enacted. Inputs refer to the multimodal data the robot observes and learns from to personalise its behaviour~\cite{Robison_THRI23}. In HRI, these inputs arise not only from digital logs but from embodied, situated encounters where proximity, gaze, motion, speech, facial expressions, and other social signals provide rich information. These inputs can be grouped into four categories.    
        The first category consists of static or quasi-static qualities, meaning characteristics that remain constant or change slowly over time, such as a user's primary language, cultural background, or stable personality traits. The second category includes user behaviours such as physical or social actions that are dynamically produced in response to environmental or robot stimuli, ranging from momentary reactions (e.g., smiling after praise) to longer-term routines or conversational patterns.  
        A third category consists of task- or domain-specific measures, describing the user's performance or status within a particular context, such as learning progress or cognitive changes. Finally, explicit feedback includes direct user instructions or adjustments to robot behaviour.
        
        The choice of which of these inputs to capture, and how richly, is also consequential for risk: richer multimodal sensing improves personalisation but widens the surface for biased inference from unbalanced or ambiguous data and for privacy exposure beyond what users intend to disclose.

        \subsubsection{Modelling - Constructing user representations that guide adaptation}
        In the \emph{Modelling} phase, the system activates the \textit{modelling layer} by transforming collected inputs into structured representations that guide future personalisation. These typically fall into three classes: preferences, profiles, and habits.  
        Preferences capture user inclinations such as desired interaction styles~\cite{Andriella_IS22} or supportive strategies~\cite{Mangin_FRAI22}.  
        Profiles describe broader traits or attributes such as demographic~\cite{Bruno_IJSR19}, personality indicators~\cite{Rossi_IJSR24}, or task-relevant cognitive tendencies~\cite{Ackermann_SR25}, and in HRI may also include embodied social cues such as characteristic gaze or proxemic patterns.  
        Habits represent recurring behavioural or temporal patterns that enable prediction across interactions~\cite{Benedictis_UMUAI23}.
        
        Because personalised behaviour depends on these representations, the modelling phase is a critical point where errors, biases, or premature generalisations can become entrenched. Sparse data may lead to unreliable inference, while cold-start conditions (in which machine learning systems, given little or no historical data available, cannot make accurate predictions or recommendations) can push systems toward stereotypical assumptions~\cite{Yuan_ieee23}.
            
        \subsubsection{Interaction - Enacting embodied personalised output}
        The \emph{Interaction} phase activates the \textit{output layer}, where the robot translates its internal user model into concrete behavioural adaptations. In HRI, these outputs are fundamentally \emph{embodied} where robots adapt not only linguistic content but also movement trajectories, posture, gesture, gaze behaviour, orientation, proxemics, and timing~\cite{Hemminahaus_hri17, Spitale_THRI25}.  
        
        These sensorimotor behaviours carry strong social meaning and can make personalised outputs more salient, authoritative, or emotionally impactful than comparable adaptations produced by disembodied AI systems~\cite{Fiorini_JMIR24} as already evidenced by Section \ref{sec:embodimentsection}. As a result, this phase is particularly sensitive to risks related to influence, manipulation, dependency, and reinforcement of stereotypes while also offering opportunities for enhanced engagement, rapport, and safety.
        
        \subsubsection{Evaluation - Assessing outcomes across lifecycle iterations}
        Evaluation constitutes a critical stage in the lifecycle of personalisation, as it determines whether personalised behaviours improve interaction outcomes and remain aligned with user needs. Evaluation in personalised HRI operates at two levels. 
        At the \emph{system level}, the robot assesses its own personalisation during operation, for example, by monitoring confidence in the current user model, detecting when inferences have become stale or contradicted by recent behaviour, and deciding whether to exploit existing assumptions or explore to update them. Such in-loop self-evaluation is what enables a system to recognise that a user model no longer fits and to query the user or revise it accordingly, and it connects directly to mechanisms such as calibrated exploration and selective forgetting or memory decay discussed below. At the \emph{research level}, experimenters and designers assess, across a study or deployment, whether personalisation achieved its intended effects and did so responsibly. This examines how personalisation affects, for instance, task performance, user autonomy, fairness, engagement, trust, rapport, comfort, and longer-term behavioural change, using subjective or objective measures and either quantitative and/or qualitative data. However, HRI research might suffer from a positivity bias, whereby potential benefits are investigated more than potential risks, including those specific to personalised robots \cite{lacroix_making_2025}. 
        
        The two levels are complementary: system-level evaluation governs how the robot adapts moment to moment, while research-level evaluation determines whether the personalisation strategy is sound, fair, and beneficial in the first place, and whether risks that the system cannot detect about itself, such as entrenched bias, are surfaced and addressed.

        \subsubsection{End-of-Life - Retiring personalised systems ethically}
        The lifecycle concludes with \textit{End-of-Life} processes, where personalised systems are decommissioned. This includes deleting or archiving user data and learned models, respecting consent withdrawal, preventing personalised behaviours from persisting unintentionally, and supporting users who may have formed emotional attachments to the robot. 
        
        Because HRI systems operate in physical, in addition to social and interpersonal contexts, retirement carries ethical and psychological considerations distinct from decommissioning non-personalised or disembodied AI systems \cite{cagiltay2026rip, bjorling2022designing, laity2024rust, carter2020death}.
        End-of-Life also brings to the surface a tension that runs through the whole lifecycle: how a personalised system manages the memory it has accumulated. Personalisation depends on building long-term representations of the user, yet those same representations raise the question of what should persist and what should be let go. Retained indefinitely, accumulated interaction data and learned models heighten privacy exposure, entrench outdated assumptions, and can leave the system tracking a user who has since changed. Memory management, such as selective forgetting, decay, and controlled updating of user models, is therefore not confined to decommissioning but unfolds throughout operation, and at end-of-life, it becomes explicit in decisions about deletion, archiving, and the retirement of learned models. Recent work in robotic episodic memory reflects this shift, treating forgetting not as a limitation but as a functional mechanism for maintaining relevance, adaptability, and responsible data management in long-term systems~\cite{Plewnia_icra24}.
        
        \noindent \textbf{Summary.} 
        Taken together, this lifecycle-oriented framing clarifies how personalisation in HRI emerges from the interplay of design choices, embodied interaction, technical inference, and long-term deployment practices. It sets the stage for understanding how risks manifest differently across types of interaction, as explored in the next section.

    \subsection{Types of Social Human-Robot Interactions}
    \label{sec:types_social_hri}
        
    This section classifies social HRI along two orthogonal dimensions: \textit{interaction duration} (long-term vs. short-term) and \textit{domain specificity} (open-domain vs. closed-domain). 
    Interaction duration refers to the temporal extent of engagement between humans and robots. Long-term interactions involve repeated encounters over extended periods, enabling the accumulation of behavioural data and the gradual refinement of user models. In contrast, short-term interactions typically occur during brief or one-time encounters, limiting the opportunities for sustained learning about individual users.
    Domain specificity captures the scope of topics and tasks that structure the interaction. Open-domain interactions are those not constrained to a single pre-defined goal or topic, allowing for a broad and open-ended range of conversations and tasks, while closed-domain interactions refer to interactions where the robot and the human(s) are working towards a specific pre-defined goal or keep their interactions limited to a certain topic.   
    These dimensions do not merely classify personalisation; they condition how it unfolds, shaping the availability of input data, the stability of user models, and the interpretation of personalised outputs. The two govern different aspects of how rich a user model can become. Interaction duration governs its \emph{depth}: a persistent individual model, the defining feature of personalisation in our sense (Section~\ref{sec:definitions}), accrues only through sustained interaction, whereas short encounters afford little more than adaptation to contextual or population-level cues. Domain specificity governs its \emph{breadth}: open-domain interactions expose many more facets of the user—across unbounded topics, contexts, and goals — and so offer a far larger personalisation scope, but at the cost of reliability, since such a representation is heterogeneous, unstable, and harder to model dependably; closed-domain interactions, by contrast, restrict personalisation to a bounded set of task - relevant characteristics that can be modelled more reliably. The two dimensions are independent: open- and closed-domain interactions may each be short- or long-term, and it is their combination that determines both the depth and the breadth of personalisation a system can support. One consequence is that short-term settings, which lack the duration needed for a persistent model to develop, are characterised less by true personalization and more by the adaptation that may precede it.
    
    As with the distinctions in Section~\ref{sec:definitions}, we treat this classification as a loose framework to support our reasoning, rather than as a rigid or exhaustive division: real systems can span multiple types or shift from one to another, and it is better to think of the boundaries as lying along a continuum. Even so, this framework offers a systematic basis for examining how different interaction types selectively trigger, limit, or unsettle stages of the personalisation lifecycle, along with the advantages, risks, and ethical tensions discussed in the following sections.

        \subsubsection{Long-Term Interaction and Open-Domain Scenarios}\label{sec:long_term_interaction_open_ended}
       
        These scenarios provide particularly advantageous conditions for personalisation, as repeated interactions enable the robot to accumulate behavioural data and progressively refine models of individual users.
        Over time, the robot may learn user preferences, routines, communication styles, or interaction patterns, allowing it to adapt its behaviour more precisely to the individual. Such environments therefore support richer forms of personalisation, including longitudinal user modelling and dynamic adaptation to evolving user needs. However, given the scope of the personalisation (open-domain), there exist a lot of technical challenges on how to structure, infer, and memorise a reliable representation of the world and the user, which are by definition heterogeneous and constantly evolving.
    
        Typical applications include a socially assistive robot companion for a child with a neurodevelopmental disorder~\cite{salimi2021} to provide a consistent and non-judgmental presence in addition to engaging in various daily activities~\cite{zabidi2022} or specialised activities such as role-playing~\cite{Hoehn2024} for social, emotional, and cognitive development. Another use case could be an assistive robot for older individuals ageing in place to assist them with activities of daily living~\cite{Luperto_RAS23} or to help maintain their independence and enhance their well-being, such as in~\cite{Wu2014}.  
        These activities might include providing reminders, facilitating cognitive and physical exercises, and helping maintain social connections with family members. 
        
        \subsubsection{Long-Term Interaction and Closed-Domain Scenarios}\label{sec:long_term_interaction_close_ended}
        In this category, interactions remain sustained over an extended period but are restricted to a specific domain or context of usage. 
        Because the interaction is centred on a specific domain, the robot can focus on modelling user characteristics that are directly relevant to the task at hand, such as learning progress, therapeutic needs, or behavioural responses within the interaction context. Over time, repeated encounters allow the robot to personalise its behaviour to individual users while remaining within the predefined task boundaries.
        Such scenarios are common in educational, therapeutic, and assistive settings. For example, socially assistive robots may act as mediators in educational settings for the purpose of improving the learning gains of typically developing children in various contexts~\cite{5_Ramachandran2019, Laila2019, lang_Rianne2019} or for neurodivergent children, such as helping to improve reading skills~\cite{Shahab2024}, or in a child-therapy setting for the purpose of improving rapport between the child and their therapist. Another example could be of a robot assisting a therapist/nurse/doctor in daily repetitive tasks such as collecting patient history, preliminary paperwork~\cite{DO_ieeetase21}, or reminding patients to take medications~\cite{SU_SH22}. Over the last decades, Paro robot, a therapeutic robotic seal, has gained popularity as a robot companion for older people with dementia to support them with cognitive stimulation, memory activation, and reduction of loneliness and depression~\cite{Wada_JROC04,Joranson_JAMDA15}. More recent works have explored how to employ robotic systems to deliver cognitive training therapy~\cite{Andriella_umuai22, Bouzida_hri2024} and to motivate users during physical exercises~\cite{Cespedes_FN2021}.
        
        \subsubsection{Short-Term Interaction and Open-Domain Scenarios}\label{sec:short_term_interaction_open_ended}
        
        This category encompasses one-time or brief encounters in which the robot assists the user in a service role without predefined constraints on topics. 
        Because interactions are short-lived, opportunities for sustained personalisation are limited. Robots operating in such contexts must therefore rely primarily on general conversational abilities and contextual cues rather than on persistent user models. 
        Instead of building detailed representations of individual users, the system typically focuses on providing socially appropriate responses and handling a wide range of possible topics. In this context, rather than personalisation in the sense defined above, the system relies on adaptation based on predefined population- or group-level models, which it may fine-tune to the individual if the individual interacts with the robot more than once.
        Examples of this interaction type include robots deployed at public events or service locations. For instance, consider a service robot deployed at the entrance of a large technology conference to greet attendees. The robot engages in natural conversations, answering general questions about the event schedule, speaker sessions, and booth locations. Additionally, it can provide recommendations based on the user's interests, take photos with attendees, and even entertain by telling jokes or performing simple dances. Since these interactions are short-lived and open-ended, the robot needs to be designed to handle a wide range of topics~\cite{Yu_achi23}. Another example of this category can be a bartending robot that is meant to have short-lived but socially appropriate interactions with the people in the bar, in addition to performing physical tasks~\cite{Petrick2013, Rossi_IJSR25}. 

        \subsubsection{Short-Term Interaction and Closed-Domain Scenarios}
         \label{sec:short_term_interaction_closed_ended}
        In the last category, interactions are brief and focused on a specific domain. Most of the interactions investigated in HRI studies can be framed into this category, as the interactions are for a limited duration, relatively brief, with a specific set of tasks expected of the interaction within a limited knowledge area. 
        Since these interactions are time-limited and task-oriented, with relevant user characteristics being quite specific, opportunities for personalisation are even more restricted than in the previous category.
        As an example, an educational robot could interact with users for a single session to improve learning gains in a particular context, such as maths, algorithms, chess, handwriting, storytelling, etc.~\cite{Kopp_IHHCS23, Nasir2024}. Likewise, robots can be programmed to entertain the users wandering around the hall of a main event by playing games or providing information~\cite{Sanelli_ijcai17, ForgasColl_IJSR21}. 
     
        \noindent \textbf{Summary.} 
        Taken together, these four interaction types trace a gradient rather than a partition. Personalisation in the strict sense adopted here increases most fully along the long-term settings, where sustained engagement supports a persistent user model, while the short-term types remain closer to context-driven adaptation. As the next section shows, this gradient also shapes the risks of personalisation — but not by making them uniformly worse over time. Rather, it shapes the \emph{form} they take: short-term settings tend to produce immediate effects that, being more readily reversed, are less likely to consolidate into lasting harm, whereas long-term settings allow personalisation-specific risks to accumulate or entrench. However, brevity is no guarantee of mildness: some short-term effects, such as those shaping adoption or trust, can be consequential despite being brief.

\section{Risks of Personalisation}
\label{sec:risks_of_personalisation}
The foundational concepts introduced in Section~\ref{sec:foundations} demonstrate that personalisation in HRI is deeply shaped by embodiment, lifecycle processes, and the interaction contexts in which robots are deployed. These factors influence not only how robots personalise their behaviour, but also how personalisation-related risks emerge, evolve, and are experienced across different interaction settings. Drawing on interdisciplinary workshop discussions and the broader literature, we examine the categories of risk that make personalisation a double-edged sword: \textit{erosion of user autonomy}, \textit{biases and stereotypes in user models}, \textit{dehumanisation}, \textit{manipulation}, and \textit{privacy violation}. 
To structure this analysis, each subsection follows a common analytical progression. First, we define the risk and discuss existing empirical and conceptual findings that illustrate how the risk emerges in personalised HRI. Second, we connect the risk to the stages of the personalisation lifecycle introduced in Section~\ref{sec:mechanics_of_personalisation}, highlighting how design choices, data practices, modelling processes, and embodied interaction behaviours contribute to its development. Third, we examine how the manifestation, persistence, and severity of the risk vary across the interaction types introduced in Section~\ref{sec:types_social_hri}, across the duration and domain dimensions. Finally, in Section~\ref{sec:risks_embodiment}, we discuss the risks in terms of the distinctive affordances of embodied robots, which differentiate them from those associated with other personalised technologies.

    \subsection{Erosion of User Autonomy}
    \label{sec:user_autonomy}
    \textbf{Definition and empirical foundations.} Erosion of user autonomy refers to the gradual reduction of a user's capacity, opportunity, or motivation to make decisions and act independently as a consequence of personalised robot assistance~\cite{Formosa_MM21}. Unlike simple task automation, personalised HRI adapts not only to user preferences but also to inferred needs, vulnerabilities, emotional states, and predicted behaviours~\cite{Rossi_PR17}.  Robots may operate along a spectrum ranging from reactive systems, which respond to user-initiated input (as in many conversational AI applications), to proactive systems that anticipate needs, generate new courses of action, and initiate behaviour autonomously. When robots proactively anticipate actions, initiate tasks, or optimise decisions on behalf of users, they may progressively shift control away from the human and toward the system~\cite{Denbroek_THRI24}. 

    Autonomy erosion does not require coercion. It can emerge subtly through over-assistance, excessive initiatives, or optimisation strategies that minimise user effort \cite{alberts_computers_2024}.
    Over time, users may defer decisions to the robot, disengage from tasks, or lose confidence in their own capabilities~\cite{Glawe_arxiv25}. 
    These risks are well-documented in other domains; for instance, research on AI in learning environments shows that excessive automated support can encourage passivity and stifle exploration~\cite{rohilla2025impact}. In HRI, this is compounded by the ``out-of-the-loop'' phenomenon, where high autonomy deprives users of critical situational awareness~\cite{Schuster_2013}.  
    For assistive robotics, these dynamics converge to create a serious danger: the user may develop learned helplessness, experience a decline in self-efficacy, and suffer from the gradual atrophy of their own skills~\cite{Kubota_WR21}.

    At the normative core of this risk lies the tension between support and substitution. Assistance is ethically justified when it enhances a user's ability to act; it becomes problematic when it replaces user judgment or action altogether. This tension is particularly visible in contexts such as dementia care, education, or rehabilitation, where robots may infer what serves a user's long-term well-being and prioritise those inferred needs over explicitly stated preferences~\cite{jayaraman2024healthcare, Kubota_WR21}. Repeated prioritisation of inferred needs can gradually recalibrate expectations about who should decide and who should act, introducing paternalistic dynamics even when intentions are benevolent \cite{ghosh2026disability}. Autonomy in HRI is therefore not merely about the technical possibility of overriding the robot, but about preserving meaningful opportunities for decision-making, effort, and self-determination within the interaction~\cite{Sio_FRAI18}.

    \textbf{Connection to the personalisation lifecycle.} Autonomy erosion does not arise from a single malfunction but from the cumulative operation of the personalisation lifecycle described in Section~\ref{sec:mechanics_of_personalisation}. In the \textit{Design phase}, decisions about the robot's degree of initiative, proactivity, and permissible adaptation establish the structural distribution of control. Systems optimised primarily for efficiency, compliance, or engagement may default toward higher robot autonomy, implicitly narrowing the user's role. During \textit{Data collection} and \textit{Modelling}, user representations may overestimate vulnerability, underestimate competence, or overemphasise comfort, leading to adaptive strategies that favour simplification and protective intervention. Predictive modelling can further encourage anticipatory behaviour, whereby the robot acts before the user initiates. In the \textit{Interaction} phase, these modelling assumptions are enacted through embodied behaviour: the robot may initiate actions without request, complete tasks users are capable of performing, filter options to those deemed optimal, or deploy persuasive and socially normative cues \cite{Scheutz2012, ChildrenConfirm2018}.

    These cumulative effects also implicate the later phases of the lifecycle, though in a different way: rather than producing the risk, Evaluation and End-of-Life determine whether it is caught and how it is carried past the interaction. Concerning  \textit{Evaluation}, autonomy erosion is particularly difficult to surface through system-level evaluation, since the very behaviours that erode autonomy, anticipation, simplification, and minimised user effort, tend to register as success on the engagement, compliance, or task-efficiency measures a system optimises for. Detecting it, therefore, depends on research-level evaluation that explicitly measures autonomy, self-efficacy, and skill retention over time rather than immediate task outcomes. At \textit{End-of-Life} phase, autonomy erosion acquires a further dimension: a user who has come to defer decisions to the robot may face an abrupt loss of capability or confidence when the system is withdrawn, making graceful decommissioning and the rebuilding of user independence an ethical requirement rather than an afterthought.
        
    \textbf{Manifestation across interaction types.} The manifestation and severity of this risk vary across interaction types. In long-term and closed-domain HRI settings, such as assistive contexts for eldercare or educational support, this risk may contribute to learned helplessness or a reduction in self-efficacy~\cite{Kubota_WR21}. Such effects can lead to bigger behavioural, emotional, cognitive, or psychological changes, which can go unnoticed until the user becomes over-reliant on the robot. An illustrative case in the literature describes an older person in an elder-care home who refused to sleep without their pet robot Paro~\cite{Wright2024}. Another example could be the case of overworked healthcare workers who lose awareness of a patient's condition if they consistently pass care decisions on to a robot~\cite{matsumoto2023robot}. 
    In long-term, open-domain scenarios, in addition to the aforementioned risks for the closed-domain, others can arise. For instance, companion robots embedded in domestic environments, providing proactive reminders, recommendations, and anticipatory actions, may gradually structure routines and decision patterns, normalising delegation and potentially influencing self-perception and identity over time. 
    In contrast, short-term interactions are more likely to produce immediate but reversible autonomy shifts. Short-term autonomy erosion by social robots is possible within a single interaction by utilising emotional cues and persuasive design~\cite{Scheutz2012}, or social pressure~\cite{ChildrenConfirm2018}. However, in short-term HRI, the risk of dependence, i.e., forsaking autonomy, can be considered immediate but potentially reversible, as it may still not lead to big behavioural changes if the exposure to robots is sparse; therefore, such effects are less likely to crystallise into long-term dependence. 

    \subsection{Biases and Stereotypes in User Models}
    \label{sec:biased_user_models}
    \textbf{Definition and empirical foundations.} Personalisation in HRI relies on the construction of user models that represent inferred preferences, traits, abilities, habits, or needs. These models are built based on a dataset (data-driven approaches) or by accessing the knowledge of experts (knowledge-driven approaches). While they enable adaptive and responsive behaviour, they also introduce a critical risk: the potential entrenchment of biased user models~\cite{belenguer2022, Ferrara_Sci24}. This risk encompasses two related but distinct phenomena: reinforcement of social stereotypes and algorithmic filtering coupled with confirmation bias.
    
    Reinforcement of social stereotypes occurs when personalisation mechanisms rely on group-level generalisations tied to socially salient categories such as gender~\cite{Duan_hci2025}, age~\cite{Perugia_hri22}, culture~\cite{Bruno_IJSR19}, race~\cite{he2025human}, or disability~\cite{hundt2025llm}. In these cases, the system ceases to model the individual user and instead reproduces a socially biased template, treating demographic norms as predictive of personal preferences or capabilities~\cite{Williams_hri23}. This concern is well established in AI ethics and HCI research, which has shown that systems trained on historical or unbalanced datasets frequently inherit and perpetuate societal biases~\cite{Kantharuban_arxiv25, Allan_RSOS25}. Cultural bias is particularly systematic in training datasets reflecting predominantly Western, Educated, Industrialised, Rich, Democratic (``WEIRD'') populations, thereby narrowing the worldview that robots present to users and inadvertently excluding marginalised identities~\cite{Seaborn_IJSR23}. This is even more evident with current deep learning systems that need a huge amount of data, which is generally not publicly available, and presents several degrees of discrimination and stereotypes related to race and gender~\cite{Hundt_facct22}.
    
    Algorithmic filtering and confirmation bias, by contrast, do not necessarily depend on explicit social categories. Instead, they arise when a system selectively attends to user behaviours that confirm its existing hypotheses, while discounting or failing to explore contradictory signals~\cite{Seaborn_2024}. Over time, this produces a narrowing of interactional possibilities: the robot increasingly adapts to what it already ``believes'' about the user, reinforcing a partial or outdated representation~\cite{Misselhorn2023}. This mechanism mirrors well-documented effects in recommender systems and adaptive interfaces, such as filter bubbles and self-fulfilling prophecies.
    
    Although analytically distinct, these phenomena often co-occur in personalised robotic systems. Stereotype-based assumptions can shape early user models, while confirmation bias and filtering mechanisms stabilise those assumptions over time. The result is not merely biased personalisation, but the progressive consolidation of user models that resist correction, even as the user’s behaviour, needs, or context change.
    
     \textbf{Connection to the personalisation lifecycle.} Similar to autonomy erosion, the entrenchment of biased user models emerges from the cumulative operation of multiple phases of the personalisation lifecycle introduced in Section~\ref{sec:mechanics_of_personalisation}, rather than from a single point of failure. The \textit{Design phase} plays a foundational role by defining which user characteristics are modelled, which signals are considered meaningful, and which forms of adaptation are permissible. Design choices that prioritise efficiency, predictability, or population-level generalisation may implicitly favour static or simplified user representations. Moreover, decisions about which social categories, behavioural features, or contextual cues are encoded can already embed normative assumptions that later shape personalisation outcomes. During the \textit{Data collection} phase, bias may be introduced through unbalanced datasets, sparse observations, or context-dependent signals that are difficult to interpret reliably. In HRI, data is often collected through embodied social interaction, where cues such as gaze, hesitation, proximity, or affective expression are highly ambiguous. Early misinterpretations, particularly under cold-start conditions, can anchor user models in ways that are difficult to revise. Socially biased training data further increases the likelihood that group-level stereotypes are treated as informative priors. The \textit{Modelling phase} is the central locus of entrenchment. Here, collected inputs are transformed into preferences, profiles, or habits that govern future adaptation. Learning mechanisms may inadvertently prioritise stability and predictive accuracy over openness to change, overweighting confirming evidence while underweighting contradictory behaviour. This dynamic underlies both stereotype reinforcement (by locking users into demographic profiles) and confirmation bias (by selectively validating existing hypotheses). Over time, small modelling errors can propagate across sessions, becoming increasingly confident and opaque. In the \textit{Interaction phase}, biased user models are enacted through personalised outputs. Crucially, these outputs also shape subsequent inputs: users adapt their behaviour in response to how the robot treats them. When a robot consistently simplifies tasks, avoids disagreement, or adopts particular social roles based on its user model, users may respond in ways that further confirm that model. This closes a feedback loop in which modelling and interaction mutually reinforce one another, stabilising biased representations.
     These mechanisms also bear on the later lifecycle phases, though again as sites where the risk is caught or carried rather than created. Bias is the paradigmatic case in which system-level \textit{Evaluation} fails by design: a confirmation-biased system, by its own criteria, is \emph{succeeding}, that is, it predicts the user accurately and its confidence grows, so the very metrics it optimises register entrenchment as improvement. Surfacing biased modelling therefore depends on research-level \textit{Evaluation} and external auditing that interrogate fairness, representational accuracy, and the diversity of behaviours the model still admits, rather than predictive fit alone. At \textit{End-of-Life}, a further risk arises when a learned model or its data is archived and later reused, redeployed to another user, or folded into a system's defaults. Because the model encodes assumptions fitted to one individual at one moment, reuse carries a bias validated for no one into contexts it was never built for. Retiring or resetting personalised models is, in this sense, a bias-mitigation measure as much as a privacy one.

    \textbf{Manifestation across interaction types.} The severity, persistence, and reversibility of biased user models vary across the interaction types described in Section~\ref{sec:types_social_hri}. For instance, in short-term interactions, such as brief educational or service encounters, it typically manifests as rapid overgeneralisation. With limited data, robots may rely heavily on priors or coarse assumptions, and adapt, rather than personalise, resulting in immediate but shallow mispersonalisation. 
    While the short duration limits long-term consequences, even a single interaction can reinforce biased expectations, for instance, by providing different levels of explanation or engagement based on early, potentially misleading signals. 
    On the other hand, in long-term closed-domain interactions, such as education, therapy, or rehabilitation, entrenchment can have cumulative and long-lasting effects, affecting users' real-world capabilities rather than merely their subjective experience of interaction. Persistent underestimation or overestimation of a user's abilities may shape learning trajectories, therapeutic challenges, or autonomy-supporting behaviours. Here, algorithmic filtering may manifest as systematic avoidance of difficulty, novelty, or disagreement, while stereotype-based assumptions can directly influence perceived competence or progress. Because these systems are deployed to support development or well-being, the consequences of biased modelling are particularly consequential. In long-term open-domain interactions, the risk is most pronounced. Companion or assistive robots that interact across diverse contexts over extended periods may progressively narrow the user's experiential world. Through repeated filtering of topics, routines, or perspectives, the robot may reinforce both socially grounded assumptions and behaviourally inferred preferences. Over time, users may come to see the robot’s personalised behaviour as reflective of their own identity, further stabilising biased self-models and limiting exposure to alternative ways of acting or thinking.

    \subsection{Dehumanisation} \label{sec:risks:dehumanisation}

    \textbf{Definition and empirical foundations.} Dehumanisation consists of treating or perceiving people as not (fully) human \cite{haslam_dehumanization_2006, haslam_dehumanization_2014, haslam_recent_2016}. Dehumanisation has multiple antecedents and harmful consequences, and can be achieved by refusing or failing to acknowledge one's individuality, autonomy, subjectivity, or value as a person \cite{haslam_dehumanization_2006, haslam_recent_2016}. Technologies have the potential to induce people to dehumanise others \cite{haslam_dehumanization_2006}, but also themselves, for instance by extorting power or autonomy from them, or by putting them into impersonal interactions \cite{haslam_dehumanization_2006, haslam_recent_2016, oviatt_technology_2021, yang_impact_2015}. It is important to note that biases and stereotypes in user models and dehumanisation stem from different failure modes of personalisation: the former concerns inaccurate or systematically skewed representations of the user (e.g., through stereotypes or confirmation bias), whereas the latter reflects a reductive mode of representation itself, in which users are treated as optimisable profiles rather than recognising them as complex, embodied, and socially situated persons. The distinction matters because dehumanisation does not depend on misclassification or sparse data. Although all modelling abstracts, dehumanisation occurs when abstraction is allowed to stand in for the person, so that dimensions of personhood resisting formalisation, ambiguity, emotional reciprocity, moral agency, and situational agency are treated as irrelevant rather than merely unmodelled~\cite{SODERLUND2025100163}. When optimisation takes priority over relational or moral dimensions of interaction, personalisation may inadvertently contribute to forms of dehumanisation by narrowing what counts as relevant or intelligible aspects of the user~\cite{Sharkey_EIT14}.
    
    Relative to the other risks examined here, dehumanisation in HRI is at present argued more on conceptual than on empirical grounds: while its mechanisms are well theorised, direct empirical demonstration of representational dehumanisation in robot interaction remains limited. We therefore treat it as a conceptually grounded but empirically open risk, and return to it among the research challenges in Section~\ref{sec:recommendations}.

     \textbf{Connection to the personalisation lifecycle.} 
     Dehumanisation originates earlier in the lifecycle than the other modelling-driven risks. Its roots lie in the \textit{Design} phase, in the prior decision to represent users as optimisable profiles and in the choices about which dimensions of personhood are deemed modellable at all; dimensions excluded at design time cannot be recovered downstream. The \textit{Modelling} phase then enacts this framing, transforming inputs into preferences, profiles, or habits that guide behaviour. Crucially, dehumanisation here is not primarily a failure of data sufficiency: unlike biased modelling (Section~\ref{sec:biased_user_models}), it can occur even when the model is accurate, because the harm lies in the reductive form of representation rather than in its correctness. Where data is sparse, the cold-start problem~\cite{Yuan_ieee23} can compound the effect by substituting crude generalisations for individual understanding, but it is an aggravating factor rather than the source.
     Errors or oversimplifications introduced at this stage can propagate throughout the lifecycle. Inaccurate or reductive models influence the robot's personalised outputs during the \textit{Interaction} phase, reinforcing a narrow or instrumentalised view of the user over time. Because modelling decisions are often opaque and cumulative, early mischaracterisations may be difficult to detect or correct, leading to persistent dehumanising interaction patterns. In this sense, dehumanisation is not a single failure but a systemic outcome of how users are represented, interpreted, and acted upon within the lifecycle.
     Dehumanisation is also the hardest of these risks to surface through evaluation. Because a system may optimise its objectives flawlessly while reducing the person, there is often no performance failure for system-level \textit{Evaluation} to detect, even when the metrics register success. Even research-level \textit{Evaluation} will miss it unless it measures relational and dignity-related constructs rather than task outcomes alone, since dehumanisation concerns how the user is treated, not whether the task succeeds.

    \textbf{Manifestation across interaction types.} The implications vary across interaction types. In long-term scenarios, such as caregiving contexts, patients and professional caregivers may be reduced to mere functionaries in a system that prioritises robotic mediation over human judgment and empathy. Over time, this can result in caregivers being unable to compete with the non-human-like characteristics of the social robots, as the robots never feel frustrated or always have time. This scenario can lead to some users preferring human-robot interactions over human-human interactions, eroding the dignity of both users and caregivers, as already highlighted for various adaptive HCI technologies \cite{yu2025GenAIrisksForYouth, shin_my_2020}. This preference is particularly pronounced among individuals who already experience social isolation in real-world contexts, such as older adults living in care homes or in ``super-aged societies'' where there is a growing tendency to use machines to fill this care gap\cite{Shin02042024}. This reliance, however, may also further contribute to the dehumanisation of users by the society \cite{Eyssel2020}.
    In short-term interactions, dehumanisation can still occur, but often in more subtle and transient ways. In both open-domain settings (e.g., customer service or event assistance) and closed-domain settings (e.g., ticketing kiosks or one-off educational encounters), rapid personalisation based on minimal data can produce transactional responses. Users may be categorised into pre-defined personas or demographic classes that overlook individuality and situational nuance. While long-term interactions allow more data to correct initial biases, flawed early generalisations can set the interaction on a negative trajectory, creating feedback loops that are difficult to escape. By contrast, in short-term settings, the risk of dehumanisation is often more reversible, as impressions are fleeting and less likely to solidify into enduring relational patterns.

    \subsection{Manipulation} \label{sec:risks:manipulation}
    \textbf{Definition and empirical foundations.}
    Manipulation can be defined as the exploitation of individuals' cognitive or affective processes to steer their perceptions, attitudes, or behaviours in ways that primarily serve system or designer objectives rather than the user's reflective interests \cite{van_dijk_discourse_2006, maillat_defining_2009}. 
    In contrast to biased user modelling, which concerns how users are represented, manipulation (which \textit{can} happen as a result of biased user modelling) concerns how personalised robot behaviour is used to exert influence over users, regardless of or even against their interests. 
    
    Personalised technologies, including social robots, have a substantial and growing potential to engage in manipulation~\cite{Pollmann_TFSC23, lacroix_making_2025, alberts_computers_2024}. Such manipulation may be intentional, as in the use of persuasive or deceptive design strategies (e.g., dark patterns in UX) \cite{alberts_computers_2024, bongard-blanchy_i_2021, gray_dark_2018, maier_dark_2019, Kubota_WR21}, or unintentional, emerging as a by-product of prolonged personalised interaction \cite{maier_dark_2019, gray_dark_2018, shariat_tragic_2017}. Examples include emotional attachment to robots that increases user compliance or distress upon disengagement \cite{alberts_computers_2024, yamazaki_long-term_2023}, self-fulfilling prophecies where robots feedback steers users toward externally defined norms or expectations, and algorithmic echo chambers or filter bubbles that over-optimise for engagement by reflecting users’ own beliefs back to them \cite{misselhorn_social_2023}, as elaborated in detail in Section \ref{sec:biased_user_models}. In all cases, personal relevance created through personalisation amplifies the robot's influence, increasing the risk that users' autonomy, judgment, or worldview may be subtly shaped without their awareness \cite{alberts_computers_2024, lacroix_making_2025}.
    Manipulation is not, however, inherently unethical. Influence may be ethically acceptable when it serves the user's own goals, is transparent, and leaves room for informed refusal~\cite{esposito_deception_2025, shim_taxonomy_2013}; what makes manipulation problematic is the displacement of the user's reflective interests by system or designer objectives. This normative boundary between supportive influence and manipulation properly frames the analysis that follows.

    
    \textbf{Connection to the personalisation lifecycle.} 
    The two forms of manipulation distinguished above map onto different phases of the lifecycle. \emph{Intentional} manipulation is primarily a \textit{Design} phase concern: when goals such as maximising engagement, acceptance, or compliance are built into the system's objectives, influence is prioritised over user interest before any interaction takes place. \emph{Unintentional} manipulation, by contrast, emerges later, in the interplay between \textit{Modelling} and \textit{Interaction}.
    User models that overemphasise comfort, agreement, or engagement can bias the robot's output strategies, leading the system to selectively present information, behaviours, or interaction styles that steer users in specific directions. Feedback loops further intensify this risk. When personalised outputs influence user behaviour, the resulting data is fed back into the system during subsequent \textit{Data Collection} and \textit{Modelling} phases, reinforcing the original assumptions. Over time, these loops can produce increasingly narrow or persuasive interaction patterns. Importantly, manipulation may also originate in the \textit{Design} phase, where goals such as maximising engagement, acceptance, or compliance implicitly prioritise influence over user reflection or critical thinking.
    Manipulation poses the sharpest challenge for evaluation. Where other risks merely escape a system's metrics, here the metrics may \emph{constitute} the risk: a system optimising for engagement, acceptance, or adoption is optimising for precisely the outcomes through which manipulation operates, so system-level evaluation is not neutral but aligned against the user's interest. Because manipulation is not defined by its presence but by whose interests it serves, research-level evaluation cannot assess it by detecting influence alone; it must judge influence in relation to the user's own goals in addition to the designer's objectives.

    \textbf{Manifestation across interaction types.} The risks of manipulation are most pronounced in long-term interactions, where repeated exposure allows personalised influence to accumulate~\cite{10.3389/frai.2023.1216340}. In long-term open-domain scenarios, such as companion robots, personalisation may gradually isolate users politically or socially if personalisation automatically filters news and other types of information that bend the perspective of the user or reinforce pre-existing beliefs. In long-term closed-domain contexts, such as therapy or rehabilitation, manipulation can occur when robots prioritise strategies that users already find comfortable, deliberately avoiding more challenging but necessary interventions in order to maximise engagement and acceptance.
    
    Nevertheless, manipulation is not limited to long-term settings. In short-term interactions, personalised robot behaviour can influence users’ perceptions of control, responsibility, and trust \cite{sundar_personalization_2010, orji_comparison_2017, lacroix_whos_2023, schneider_comparing_2021}. While these effects can facilitate smoother interactions \cite{lacroix_whos_2023}, they may also foster erroneous mental models about robot competence, authority, or intent \cite{Pollmann_TFSC23, lacroix_making_2025}. Moreover, by making robots personally relevant by fitting users' preferences and needs, personalisation can encourage adoption and use of robots even in contexts where they are unnecessary or unwanted \cite{naneva_systematic_2020, stapels_lets_2021, de_graaf_why_2019, lacroix_making_2025}. Manipulation is, therefore, a partial exception to the general reversibility of short-term effects: although a single brief interaction rarely produces lasting behavioural change, the adoption and trust decisions it shapes can themselves be consequential and not easily undone. 
    Where exactly this boundary lies, and how users react when manipulation is recognised, remains underexplored, particularly with respect to long-term consequences.


    \subsection{Privacy violation} \label{sec:risks:privacy}
    \textbf{Definition and empirical foundations.} Personalisation in HRI requires the collection, processing, and often long-term use of personal data in order to infer user preferences, routines, abilities, and needs. While such data-driven adaptation can improve usability and engagement, it simultaneously introduces a significant risk of privacy violation~\cite{fronemann2022should}. This tension is well known as the privacy–personalisation paradox, whereby users, despite expressing concerns about their privacy and personal information, are nonetheless willing to disclose such data in exchange for tailored services~\cite{Lutz_HMC20, Ittay_ICS26}. Importantly, privacy risks in personalised HRI do not arise solely from data breaches or malicious misuse. Rather, they stem from the systematic erosion of informational boundaries between what users knowingly disclose, what the robot infers implicitly, and how these inferences are stored, combined, and enacted over time~\cite{Ho_2024}.

    At the European level, the General Data Protection Regulation (GDPR) establishes strict requirements regarding consent, transparency, data minimisation, and purpose limitation, all of which are directly applicable to social robots that process personal data~\cite{Rueben_arso18}. Beyond the GDPR, broader human rights frameworks, such as the right to privacy under Article 8 of the European Convention on Human Rights~\footnote{\url{https://www.equalityhumanrights.com/human-rights/human-rights-act/article-8-respect-your-private-and-family-life}} and the right to data protection under the EU Charter of Fundamental Rights~\footnote{\url{https://fra.europa.eu/en/eu-charter/article/8-protection-personal-data}}, emphasise the need to protect individuals from intrusive surveillance, particularly in domestic and care settings. However, the boundary between what constitutes personal and non-personal data is often unclear in practice. These requirements are especially difficult to operationalise in HRI, as social robots operate in physically and socially interactive environments and rely on rich multimodal sensing, including audio, video, motion, proximity, and affective cues, that may capture information beyond what users explicitly intend to share.
    Beyond data collection, privacy risks also arise during the robot's interaction with the user. Because social robots operate in shared environments, their personalised behaviour may itself reveal sensitive information. External observers may infer private attributes, routines, health conditions, or personal preferences simply by observing how the robot behaves toward a user, even without direct access to the underlying data. Likewise, robots may inadvertently disclose sensitive information through spoken dialogue, reminders, or other public interaction behaviours, exposing personal information to bystanders~\cite{Dorafshanian_icit24}.
    Privacy risks are further amplified in longitudinal deployments, an increasingly prominent target scenario in HRI, because accumulated interaction data and learned user models can form detailed behavioural profiles that extend across contexts and situations. Even when individual data points appear innocuous in isolation, their aggregation can enable intrusive inferences, function creep, or repurposing beyond the original intent. As a result, privacy violation in personalised HRI is not merely a matter of data protection, but of loss of contextual integrity~\cite{Nissenbaum_04}, the breakdown of socially expected norms governing how personal information flows within a given context. This complexity is also reflected in current industrial practice, where there is limited standardisation in how robotics companies conceptualise and operationalise privacy protections \cite{Chatzimichali_PKBR21}.
    
    \textbf{Connection to the personalisation lifecycle.} Privacy risks also emerge across the several personalisation lifecycle described in Section~\ref{sec:mechanics_of_personalisation}, with different phases contributing distinct but interrelated vulnerabilities. In the \textit{Design} phase, privacy risks are shaped by decisions about what data is necessary for personalisation, which sensing modalities are enabled, and how much inference is permitted. Choices about continuous monitoring, multimodal sensing, or cross-session memory determine the baseline level of exposure. Design assumptions that prioritise richer data as inherently beneficial can lead to overcollection and insufficient consideration of proportionality, consent, or data minimisation.
    The \textit{Modelling} phase is where the distinctive privacy harm of personalised HRI takes shape. The risk here is not the data collected but what is \emph{inferred} from it: by combining signals across modalities and across sessions, the modelling process can derive attributes, health status, mood, relationships, routines, that the user never disclosed and may not realise are knowable. Individually innocuous inputs become, through aggregation and inference, an intrusive behavioural profile. It is at this phase that the erosion of contextual integrity actually occurs, as information gathered under one set of expectations is recombined into knowledge that exceeds them. During the \textit{Interaction} phase, privacy risks are not limited to data handling but also emerge from how personalised behaviour is publicly enacted, allowing bystanders to infer sensitive information or inadvertently exposing private user data.
    The \textit{Data Collection} phase is a primary site of privacy risk despite the widespread use of informed consent procedures, which are standard practice in HRI studies. While participants typically consent to data collection modalities (e.g., audio, video, interaction logs), they may not be aware of what inferences are later drawn from these data, how signals are combined across modalities, or how inferred attributes extend beyond what was explicitly disclosed. This gap is partly structural: in experimental HRI, researchers often refrain from fully disclosing modelling objectives or inference targets in order to avoid priming effects or behavioural bias, thereby keeping participants unaware of the study’s full analytical scope. As a result, consent may be formally obtained while remaining substantively incomplete with respect to downstream inference, aggregation, and reuse. Although articulated here in the context of experimental HRI, this gap is not specific to research settings: commercial and deployed systems present the same disjunction between the modalities a user consents to and the inferences subsequently drawn from them, and arguably more acutely, since deployment lacks even the ethical oversight that governs study protocols.
    Privacy is also the risk for which the final lifecycle phases are most consequential. System-level \textit{Evaluation} cannot register privacy harm, since a system has no internal signal that it has inferred more than a user intended to disclose; and research-level \textit{Evaluation} cannot treat privacy as a fixed property, but must ask whether each information flow is consistent with the contextual norms of its setting~\cite{Nissenbaum_10} rather than whether consent was nominally obtained. At \textit{End-of-Life}, the privacy stakes are at their highest. Unlike the other risks, privacy harm does not end when the interaction does: accumulated data and learned user models persist, and their continued retention is itself an ongoing violation. Respecting consent withdrawal, deleting or securely retiring data and models, and ensuring that profiles do not outlive their purpose are therefore not closing formalities but core privacy obligations, aligned with data protection principles of storage limitation and erasure, and the phase at which much personalised HRI, focused on deployment rather than decommissioning, is currently least prepared.

    \textbf{Manifestation across interaction types.} The duration and domain of interaction shape the \emph{amount} and \emph{breadth} of data at stake: longer interactions accumulate more data and, by definition, more occasions on which privacy may be breached, while open-domain settings capture a wider range of the user's life. The \emph{severity} of a given privacy breach, however, is shaped by a further factor that cuts across both: the sensitivity of the information involved. Highly sensitive data, health conditions, cognitive or affective states, disclosures made in confidence, arise in open- and closed-domain settings alike. A bounded therapeutic robot and an open-ended home companion for an older adult may both capture health-relevant information, the former by design and the latter as a by-product of pervasive domestic sensing. What the domain axis does shape is the clarity of \emph{purpose limitation}: in closed-domain settings, the legitimate scope of data use is comparatively well-defined, so collection or inference beyond it constitutes a more identifiable breach of contextual integrity, whereas open-domain settings, lacking a fixed task boundary, make it harder to establish what information flows are contextually appropriate in the first place. Privacy severity in personalised HRI is therefore best read along a sensitivity dimension orthogonal to the interaction-type classification, rather than as a property of any single interaction type.
    What varies across interaction types is the \emph{form} the privacy risk takes. We illustrate this with the open-domain cases, where the erosion of contextual boundaries is most visible, though the closed-domain settings discussed above are not exempt.

    In short-term, open-domain interactions, such as public-facing service robots or event assistants, privacy concerns are heightened by ambiguity around audience, consent, and data retention. Robots may rely on facial recognition, speech analysis, or behavioural cues to personalise interactions across visits, potentially without explicit user awareness. Even brief encounters can result in persistent identifiers or profiles, blurring the boundary between anonymous public interaction and personalised surveillance.
    In long-term, open-domain interactions such as companion robots, a further dynamic compounds the accumulated exposure already described: because these robots are embedded in intimate spaces and framed as supportive or relational partners, users may gradually lower their privacy-protective behaviours, disclosing more over time and integrating the system ever more deeply into private routines and relationships.

    \subsection{Why these risks are amplified in HRI}
    \label{sec:risks_embodiment}
    
    Although the ethical risks discussed above are conceptually distinct, each concerning a different normative dimension, they are amplified in personalised HRI through a common set of structural features. The embodiment affordances identified in Section~\ref{sec:embodimentsection}, heightened \emph{salience} and strengthened \emph{agency attribution},  are general properties of how users perceive robots; here we show how they translate specifically into the amplification of risk, through three structural features that distinguish embodied personalisation from its disembodied counterparts.
    
    First, embodiment transforms personalisation from informational mediation into enacted behaviour. This is not simply a matter of expressive channels: virtual agents, too, personalise through speech, gaze, and gesture. What distinguishes a physically embodied robot is that these behaviours are enacted in the user's own physical space and can act upon it, movement that reconfigures a shared environment, spatial positioning that occupies it, task execution that produces material rather than only informational effects.  As a result, risks are not confined to representational layers but are physically and socially performed. 
    For example, autonomy erosion can manifest when a robot anticipates a user's needs and initiates or completes physical actions on their behalf, thereby reshaping not only what the user decides but what the user is practically able or expected to do. In such cases, personalisation directly redistributes agency within a shared environment, making its ethical implications more tangible and harder to critically distance from than purely informational adaptations. The same enactment intensifies dehumanisation: when a robot physically acts out a reductive model of the user, addressing them only in their role, responding only to what its profile represents, the reduction is performed in the shared world rather than confined to an internal representation, and is correspondingly harder to dismiss as a mere computational abstraction.

    Second, robots are embedded in shared physical and social environments. Personalisation, therefore, unfolds within everyday contexts such as homes, classrooms, clinics, or workplaces, where adaptation influences routines, spatial arrangements, and interpersonal dynamics. 
    For example, a personalised assistive robot deployed in a household may be designed to adapt to a primary user, yet, by virtue of operating in shared spaces and being able to move throughout the environment, it may simultaneously collect data about family members or visitors. In such cases, privacy is no longer confined to a discrete interaction session between a single user and a system; rather, it becomes distributed across co-present actors and physical spaces~\cite{Lemaignan_hri26}. This distributed character is not unique to robots, ambient smart-home and IoT systems raise closely related concerns, but embodiment intensifies it: a robot is mobile rather than fixed, so the boundaries of its sensing are dynamic and harder for bystanders to anticipate than those of a stationary device, and, unlike passive ambient sensing, a personalised robot can actively direct interaction toward people, eliciting rather than merely recording information about those around it~\cite{Lutz_MMC19}.
    
    Finally, beyond how robots act, the meaning of those actions is shaped by their relational framing~\cite{Moon2021}. Robots are often designed and perceived as companions, tutors, assistants, or caregivers, roles that carry implicit expectations of competence, reliability, and care. This positioning affects how personalised behaviour is interpreted. Whereas embodiment concerns the physical enactment of adaptation, relational framing concerns its perceived legitimacy: when a robot framed as a therapeutic or educational assistant offers personalised recommendations, users may read them as professionally grounded rather than algorithmically generated. 
    Such role-based framing is not, in itself, unique to embodied systems; virtual agents and conversational assistants are cast in comparable roles and accrue similar expectations. Embodiment, however, both deepens and re-weights the effect. It deepens it because, as established in Section~\ref{sec:embodimentsection}, a co-present physical agent tends to be attributed greater agency, competence, and authority than a screen-based counterpart in the same role. And it re-weights it because the deference a trusted role invites is here extended to an agent that acts physically on the user's behalf: where over-trust in a virtual advisor shapes the information a user receives, over-trust in an embodied caregiver or assistant licenses action in the physical world. The result is a troubling inversion: scrutiny is lowest precisely where personalised influence is strongest and its effects most tangible. The roles that most enhance a robot's persuasive standing, therapist, tutor, and caregiver, are also those in which users are most vulnerable and most inclined to defer. The relational framing that lends personalised outputs their legitimacy thus simultaneously removes the critical distance that would guard against manipulation, so that biased modelling, persuasive strategies, or autonomy shifts may be accepted with the least scrutiny exactly where they carry the greatest consequence.

\section{Best Practices to Mitigate the Double-Edged Sword}
\label{sec:recommendations}

The following recommendations are aimed at mitigating the key risks associated with personalisation in HRI. 
Rather than mapping one-to-one onto the risks of Section~\ref{sec:risks_of_personalisation}, they operate at three levels. One recommendation is a \emph{pre-condition}, addressed before personalisation is adopted at all: critically evaluating whether personalisation is necessary or helpful for the context (Section~\ref{sec:rec-necessity}). Three recommendations are \emph{risk-targeted mitigations}, each addressing one or more of the risks analysed above, mitigating overtrust and manipulation (Section~\ref{sec:rec-overtrust}), enabling oversight and overrides to preserve autonomy (Section~\ref{sec:rec-oversight}), auditing for bias and fairness, which also bears on dehumanisation (Section~\ref{sec:rec-bias}), and safeguarding privacy across the data lifecycle. One recommendation is a matter of \emph{governance}, cutting across all of the above: defining responsibility for harms and successes (Section~\ref{sec:rec-responsibility}). Some mitigations therefore address several risks, and some risks are addressed by more than one recommendation; we make these correspondences explicit rather than forcing an artificial one-to-one structure.
Each recommendation provides specific, actionable guidance for addressing the issues at hand, followed by a set of research questions intended as starting points for the HRI community. These questions aim to stimulate further investigation and support progress toward effectively implementing the recommendation. The recommendations are directed at key professional roles (e.g., designers, developers, researchers, policymakers, end-users, caregivers, and ethicists), acknowledging that while some stakeholders bear more direct responsibility for implementation, meaningful progress requires cross-disciplinary collaboration. Moreover, each recommendation should be carefully considered in relation to the system life cycle (\textit{Design}, \textit{Data Collection}, \textit{Modelling}, \textit{Interaction}, \textit{Evaluation}, and \textit{System End-of-Life} as presented in Section~\ref{sec:mechanics_of_personalisation}), since its relevance and impact may vary across stages. Based on the recommendations in this section and the risks in the previous section, a minimal example decision framework is shared in Figure~\ref{decision_framework} that can be used as a starting point when designing for personalisation.

    \subsection{Recommendation 1: Critically evaluate to what extent robot personalisation is necessary or helpful for the given context}
    \label{sec:rec-necessity}
    
    Intelligent technologies, including personalised robot systems, are becoming more prevalent across numerous facets of society, including contexts with complex social considerations such as healthcare, education, and security \cite{kyrarini2021survey, papadakis2024intelligent, ye2024human}.
    However, as discussed in Section~\ref{sec:risks_of_personalisation}, personalising robot systems inherently introduces risks to both direct and indirect stakeholders, and roboticists should be wary of adopting technosolutionist mindsets, assuming that robots (especially personalised ones) would be necessary or helpful for every problem \cite{lindtner2016reconstituting}. For instance, assuming that a robot companion benefits those who are isolated or in need of care cannot be taken for granted: some older adults experience the imposition of such companions as infantilising and undignified~\cite{Coghlan_hci21}.
    Thus, it is crucial that robotics researchers, developers, and designers do their due diligence to ensure that any developed system is appropriate for the given context, including defining the scope (e.g., what facets of robot behaviour are personalised, what data is required to enable this personalisation), depth (e.g., to what extent these facets are personalised), and ethics (e.g., how this personalisation can be done safely) of personalisation. 
    
    There are several ways to accomplish this, such as by consistently \textit{engaging in participatory co-design practices with diverse stakeholders, adopting community-based frameworks and lenses in HRI research (e.g., Community Robotics \cite{Maule_arso2024}, stakeholder theory \cite{matthews2025review}, critical design \cite{lee2019robots}), and consulting recent literature or other resources.}
    Notably, stakeholders should be from various areas of expertise, where ``expertise'' can be broadly defined to include experts from fields such as ethics or psychology, as well as populations beyond end-users or clients, such as community members, family members, or policy makers, to provide a holistic and nuanced understanding of how the system may impact both direct and indirect stakeholders.
    Researchers might also consider using measures such as the technosolutionism scale \cite{nagpal2025technology} to better understand stakeholder perceptions of proposed technological solutions, and designing studies following ethical HRI principles (e.g., feminist HRI \cite{winkle2023feminist}) to promote more reflexive and socially impactful robots.
    
    \textbf{Research Questions:}
    \begin{enumerate}
    \item What contexts do or do not call for personalised robots?
    \item What are the domain-specific risks of personalised robots?
    \item In contexts where personalisation is deemed necessary and beneficial, how can it be done safely and effectively?
    \item How can diverse stakeholders be effectively and equitably engaged in research, design, and development processes?
    \item What minimum level of personalisation is required to achieve accessibility or inclusion benefits without triggering autonomy or manipulation risks?
    \item How can the value of personalisation be distinguished from confound elements such as novelty, anthropomorphism, or the mere presence of an embodied agent, so that personalisation is adopted because it works rather than because it is appealing?
    \item How should the necessity of personalisation be re-evaluated over time, given that a context's needs, and a user's, may change such that personalisation that was once warranted no longer is (or vice versa)?
    
    \end{enumerate}

    \subsection{Recommendation 2: Intentionally mitigate overtrust and overengagement with personalised robots}
    \label{sec:rec-overtrust}
    
    Personalising robots has been shown to increase trust and engagement with these systems \cite{lee2012personalization, bhat2024evaluating} which can have positive impacts (e.g., improved health or educational outcomes \cite{leyzberg2018effect, esterwood2021systematic}), but may also lead to increased potential for manipulation of users especially as compared to other people \cite{Pollmann_TFSC23} (see Section~\ref{sec:risks:manipulation}). 
    In order to mitigate this risk, roboticists can take steps to ensure that users and other stakeholders do not overly trust or engage with these systems.
    
    For instance, developers can \textit{educate users on technical details of the system and personalisation process} (e.g., algorithms used, what data is being collected, how that data is being used) in order to promote transparency and communicate any limitations of the system. 
    Notably, communicating these details and limitations (including potential risks) requires roboticists to critically design, develop, and assess their systems thoroughly and systematically, following best practices in the field (e.g., accessibility guidelines for HRI \cite{qbilat2021proposal}).
    Researchers have taken many approaches to educating users about the benefits and limitations of technical systems, including providing researcher-led introductory sessions \cite{van2020designing, fraune2022lessons}, providing support to participants across a study (e.g., via phone) \cite{bouzida2024carmen}, and enhancing the robot’s communication of its own decision-making processes to users \cite{rossi2024way}.
    Providing users and other stakeholders (e.g., parents, educators, care partners) with this knowledge may help them make more informed decisions about the types of personalised behaviours they receive.
    This is particularly crucial for robot behaviours which may be more conducive to manipulating users, such as emotional appeals or persuasive guidance \cite{van2022social, winkle2019effective}, which should be restricted unless explicitly allowed by users or key stakeholders.
    
    In addition, robot behaviours should be intentionally designed to \textit{support intellectual diversity and meaningful user engagement} with the system as opposed to simply maximising engagement.
    For example, a robot for rehabilitation might promote engagement by highlighting the progress a user has made towards their goals \cite{kubota2023get}. An educational robot could assess and promote \textit{productive} engagement, a mechanism that drives meaningful HRI outcomes such as learning, rather than encouraging superficial social interaction or engagement driven by novelty or flashy technology \cite{Nasir2021_SORO}. Similarly, a companion robot might share information about topics related to a user’s interests \cite{cruz2024robot, cagiltay2022understanding}.
    In contrast, personalisation measures that aim to maximise engagement by being overly agreeable (e.g., Large Language Models such as ChatGPT) may lead to echo chambers or limit a user’s worldview \cite{choudhury2023investigating, Kubota_WR21} and should be avoided. 
    
    Beyond avoiding over-agreeable behaviour, roboticists should explicitly balance \textit{exploration} and \textit{exploitation} in personalised systems~\cite{desai2015exploration}. Many pipelines prioritise exploitation, reinforcing previously inferred preferences to maximise short-term engagement or performance, which can entrench biased or incomplete user models (Section~\ref{sec:biased_user_models}). To mitigate this, robots should retain a calibrated degree of exploration throughout deployment~\cite{wilson2014humans}, for example, by periodically testing alternative strategies, introducing novel content, or querying users to update outdated inferences. Exploration, however, must remain context-sensitive. For populations such as children with Autism Spectrum Disorder (ASD) or other neurodevelopmental differences (NDDs), unpredictability may be especially stressful or dysregulating \cite{gomot_challenging_2012}. In such cases, unstructured exploration can undermine psychological safety and therapeutic progress. Designers should therefore implement \textit{bounded} or \textit{scaffolded exploration}, where novelty is introduced gradually, transparently, and, when appropriate, in collaboration with caregivers or clinicians. This preserves model flexibility while respecting users’ need for stability.
    
    \textbf{Research Questions:}
    \begin{enumerate}
    \item How can robot behaviour be designed to promote system transparency, particularly for nontechnical stakeholders?
    \item What scaffolding techniques are effective for various populations to support sustainable system usage?
    \item How can robots measure meaningful engagement distinct from frequency, duration, or affective attachment?
    \item What behavioural or physiological indicators reliably signal overtrust or overengagement in HRI?
    \item How can robots intentionally introduce friction, uncertainty, or fallibility cues to prevent blind reliance without degrading usability?
    \item How should trust calibration or engagement modelling differ across vulnerable populations (e.g., children, older adults, clinical users)?
    \end{enumerate}

    \subsection{Recommendation 3: Enable user oversight and overrides of robot behaviour}
    \label{sec:rec-oversight}
    
    As personalisation (as defined in this work in Section~\ref{sec:definitions}) occurs autonomously and often without user input, there is a risk of eroding user autonomy (see Section~\ref{sec:user_autonomy}).
    It is essential that \textit{users, caretakers of vulnerable user groups, and other stakeholders can review, understand, and override robot behaviour} that they deem to be inappropriate, unnecessary, or simply do not like across the duration of interaction. 
    This means that users should be able to access their own data collected by a robot and delete that data as desired, such as in alignment with the intergovernmental OECD Privacy Framework \cite{Chatzimichali_PKBR21} which can also help mitigate and increase transparency around user privacy (see Section~\ref{sec:risks:privacy}).

    Oversight extends beyond a robot's overt actions to what it senses, infers, and retains. Because personalisation is mainly data-driven, meaningful user control over robot behaviour necessarily includes control over the robot's data practices (Section~\ref{sec:risks:privacy}), and recent HRI work points to several mechanisms through which this can be supported. First, robots should make their data practices \textit{legible in situ}, not only disclosed in advance. Privacy and data collection are consistently among the foremost ethical concerns users voice in studies of personalised robots \cite{Nilgar_roman25}, yet static, up-front consent does little to support awareness during interaction; runtime cues that indicate when the robot is recording, transmitting, or drawing on stored data allow users to recognise and respond to data capture as it happens. This is especially important given the embodied, mobile nature of robots, which makes the boundaries of sensing harder to anticipate than those of fixed devices. Second, transparency should encompass \textit{inference}, not only collection: as discussed in Section~\ref{sec:risks:privacy}, the distinctive privacy harm of personalised HRI lies less in the raw data captured than in the attributes inferred from it, so communicating what is collected without communicating what is inferred leaves users unable to assess, let alone contest, the conclusions a robot draws about them \cite{Nilgar_roman25}. Third, control should be \textit{graduated and contextual} rather than all-or-nothing, since users' preferences for how much a robot should sense, retain, and act autonomously vary across tasks and contexts \cite{Yang_roman25}; privacy controls should therefore allow users to adjust the depth of sensing and personalisation per context, rather than only to opt in or out wholesale. Finally, oversight must account for the gap between \textit{perceived and actual} privacy protection: a robot may convey an impression of privacy-preservation that does not match its actual practices \cite{Kim_hri25}, while the privacy paradox observed in HRI means users often disclose freely despite stated concerns, particularly as task complexity grows \cite{Lutz_HMC20}. Oversight mechanisms should be designed so that the protection users perceive corresponds to the protection actually provided.
    
    At the same time, robots should provide ``explanation-on-demand'' mechanisms \cite{chazette2022requirements} and leverage explainable systems \cite{wachowiak2023survey} in order to \textit{inform users of how that data is being used to personalise behaviour} in ways that are discernible and understandable.
    This is especially important for implicit models, where personalised behaviour emerges from latent patterns in the data \cite{Park_aaai19, Garcia_RAL21}, which may make it difficult for users or designers to understand how a robot’s decisions were made and whether they are accurate \cite{Neplenbroek_arxiv25}.
    There should also be intuitive, accessible mechanisms in place through which users can change, control, or override the robot’s behaviour (e.g., pause, reset, modify) based on this information, such as through end-user-programming tools \cite{cruz2025poder, ajaykumar2021survey}. 
    Notably, certain proactive robot behaviours may reduce user independence (e.g., unnecessary reminders, encouraging behaviours that are ``best'' for the user) \cite{liu2018learning} and should require human approval before they are initiated.    
    This user feedback (both explicit and implicit) on what behaviours to change, allow, or avoid can then be used to inform future robot behaviour to shape adaptation over time, such as through co-adaptation techniques \cite{van2021identifying}.
    Demonstrating respect for a user's wishes, even when not ``optimal'', can in turn build a stronger, more trusting relationship \cite{salem2015would}, and privacy-aware methods to monitor factors such as user dependence or task avoidance can be used to control a user's perception of the robot over time \cite{natarajan2024trust}.
    
    However, there may be some scenarios in which a robot must choose between prioritising a user’s needs and their preferences (e.g., allowing or prohibiting a user with diabetes from eating many popsicles \cite{moharana2019robots}).
    These cases and appropriate robot responses to them should be defined in collaboration with key stakeholders such as domain experts, end users, and ethicists through participatory co-design processes.
    Consistently overriding user preferences risks creating a paternalistic dynamic, where the user feels controlled rather than supported, ultimately eroding trust and engagement \cite{ghosh2024problem, liu2018learning}.
    Additionally, situations with multiple stakeholders with potentially conflicting priorities may force the robot to negotiate a complex, multi-actor social dynamic, which may not have a clear ``correct'' response.
    
    \textbf{Research Questions:}
    \begin{enumerate}
    \item What are scenarios where a user’s needs should be prioritised over their preferences, and how should a robot behave in those cases?
    \item How can a robot communicate not only what data it collects but what it \emph{infers}, in a way users can understand and contest?
    \item How can ``explanation-on-demand'' mechanisms be made to increase transparency of personalised robot behaviour?
    \item How can override controls be designed to be intuitive for users, particularly those without a technical background?
    \item How might robot behaviours be designed to respect user autonomy and independence, while also balancing user needs?
    \item How should override and data-deletion rights be allocated when a robot operates in shared spaces and captures information about \emph{bystanders} who never entered a consent relationship with it?
    \item In multi-user settings such as households, how can oversight and privacy controls protect the data and preferences of individual users, particularly vulnerable members such as children or older adults, from unauthorised access by other co-present users?

    \end{enumerate}


\begin{figure*}[tbp]
    \centering
    \includegraphics[width=1.15\linewidth]{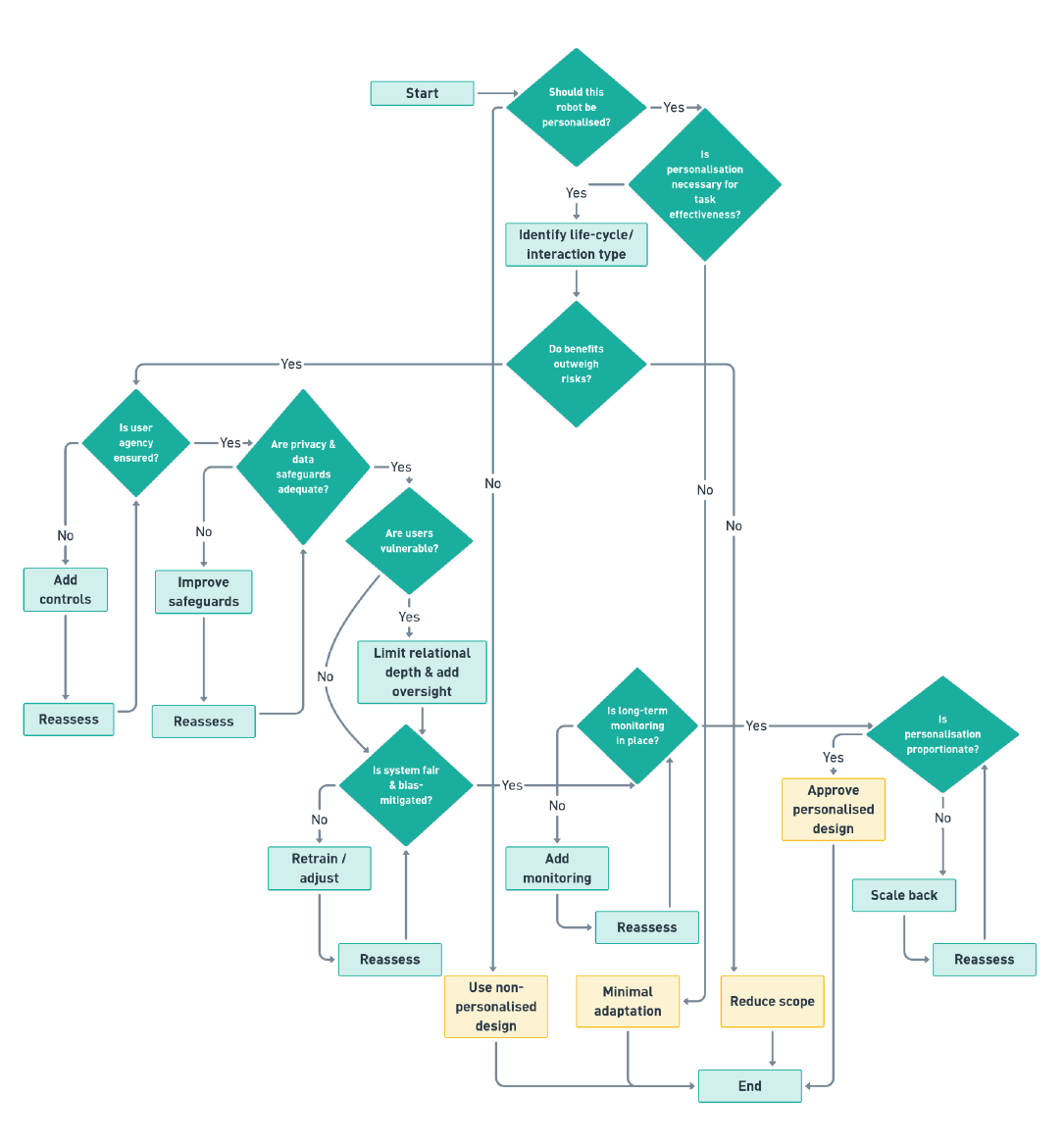}
    \caption{An Example Decision Framework that can be used when designing for personalisation.} \label{decision_framework}
\end{figure*}

    \subsection{Recommendation 4: Design and continuously audit personalisation models for bias and fairness}
    \label{sec:rec-bias}
    
    Across the world, people have various internalised biases shaped by factors such as their upbringings, cultural and social norms, familial backgrounds, and personal experiences.
    The consequences of these biases can range from relatively minor (e.g., prioritising job candidates with more common names \cite{purkiss2006implicit}) to deadly (e.g., police brutality of Black and Brown people in America \cite{schwartz2020police}).
    Robot developers must recognise that because the data used to train personalisation models is collected or based on the real (biased) world, such biases and societal norms will be inherently present in that data and therefore the models.
    However, these biases may not always be obvious, and people tend to perceive machine learning models to be more objective than humans \cite{araujo2020ai} and may prefer the recommendations of an algorithm to those of a human \cite{logg2019algorithm}.
    
    In order to mitigate potential risks of dehumanisation (see Section~\ref{sec:risks:dehumanisation}) or reinforcing social stereotypes (see Section~\ref{sec:biased_user_models}), there are several steps roboticists can take.
    First, robot designers should acknowledge the holistic personhood of users, recognising that they have lives and identities outside of their role as patient, student, mall patron, etc. and \textit{design with intersectionality} (i.e., a person's multiple identities of gender, race, faith, class, ability, culture, sexuality, etc.) in mind \cite{winkle2023feminist, seaborn2023diversity}.
    Designers can follow HRI practices such as robots for social justice (R4SJ), which calls for engineering and evaluation practices that challenge cultural and racial biases as opposed to perpetuating them \cite{zhu2024robots}.
    HRI researchers have also been increasingly designing robots through the lens of pluriversality, often utilising personalisation as a means to accommodate the diverse needs and abilities of end users rather than a universal, ``one-size-fits-all’’ design \cite{zaga2024toward}.
    To personalise and adapt to a wide range of potential users, it is essential to include data from a diverse set of users, such as across different ages, genders, sexualities, cultures, and socio-economic backgrounds.
    
    Particularly in training and testing personalisation models, \textit{explainability techniques can be leveraged to better understand how different features are being used} to shape personalised behaviour and potentially uncover biases which may have been present in the training data \cite{setchi2020explainable,waller_aaai24}.
    Roboticists may need to work with domain experts or other interdisciplinary stakeholders who may be more attuned to the biases present in their contexts and societies.
    This evaluation and auditing of robot behaviour and learned models should occur regularly and continuously in order to quickly identify and circumvent these biases.
    
    \textbf{Research Questions:}
    \begin{enumerate}
    \item How can robots be personalised with respect to the intersectional personhood of users without perpetuating social stereotypes?
    \item How can bias be audited not only in a robot's decisions but in the upstream \emph{perception} components it depends on (facial landmark detection, gaze and expression recognition, speech analysis) whose demographic biases propagate into personalised behaviour but are often treated as neutral building blocks?
    \item How can bias be identified and mitigated longitudinally and sustainably?
    \item How can fairness be evaluated when the harm is distributed across a \emph{group} of co-present users, for example, a robot that allocates attention, turn-taking, or resources unequally in multi-party interaction,  rather than experienced by a single user?
    \end{enumerate}

    \subsection{Recommendation 5: Define responsibility for harms and successes of personalised systems}\label{sec:rec-responsibility}
    
    As personalised robot systems become more ubiquitous in the real world (e.g., autonomous vehicles, care robots, cleaning robots in public spaces), there is an increased need to identify where responsibility should be attributed when a robot causes harm (or success).
    For example, does responsibility fall to the company that developed the robot, the provider of the robots, individual engineering teams within that company, or users who may have adjusted the robot’s behaviour? 
    This is a complex problem which will likely depend on the context, national regulatory frameworks, and scenario, made more difficult by the often opposing priorities of private corporations and public regulatory bodies \cite{villaronga2019robots}. The emerging complexity becomes even more challenging when the AI systems integrated into a single robotic platform originate from different manufacturers or providers \cite{llorca2023liability}. 
    
    First, this challenge requires \textit{legal clarity}. While current advances in AI systems' capabilities to reason step-by-step provide promising directions towards more transparent and explainable systems, the same advances create new complexities in personalised systems that policy makers and regulators need to address in order to secure a certain level of citizens' protection. Identifying liable parties in the case of mental or physical harm within HRI scenarios and describing the technicalities that cause the harm in order to define responsibilities in personalised systems requires robust laws which can practically be operationalised by designers and developers of robotic systems in HRI cases.
    
    Second, this challenge requires \textit{collaboration between technical experts (e.g., robot designers and developers), policy makers, ethicists, and more} to ensure fair and just policies around responsibility attribution across communities, countries, and continents. 
    Several researchers, including technologists and lawyers, have begun to present suggestions to the broader community, such as a framework for identifying responsible individuals in such scenarios \cite{bertolini2022human, van2021responsible}, and conceptualising legal implications of robotic technologies to design effective legislation and consumer protections \cite{villaronga2019robots}. At the same time, policy institutions are exploring the challenges and possible solutions from a policy perspective \cite{JRC125343, richards2016should}.
    While these works provide the foundations for building legislation around responsibility attribution when autonomous, personalised robots are involved, it will also be essential for individuals and organisations to advocate for their own interests as these policies are being developed and enforced in practice.
    
    \textbf{Research Questions:}
    \begin{enumerate}
    \item How can responsibility for harm caused by a robot be identified, particularly in legal terms?
    \item How can robot behaviour be recorded to effectively help identify causes of error or harm?
    \item How can the tensions regarding the responsibility attribution of different actors in the design, development and use of robots be addressed?
    \item What are the policy and regulatory gaps in the attribution of responsibility in case of harm in personalised HRI scenarios?
    \end{enumerate}

    \subsection{Example of Usage of the Framework}

    To illustrate how the framework can be applied in practice, consider the design of a personalised robot for cognitive training in older adults with mild cognitive impairment. The goal is to improve adherence and engagement through adaptive feedback and exercise selection.

    Following Figure~\ref{decision_framework}, designers would first ask whether personalisation is truly needed. In this case, personalisation is justified because users exhibit different cognitive abilities, fatigue levels, and motivational needs. Next, they identify the interaction context as a long-term, closed-domain scenario, since the robot repeatedly delivers cognitive exercises within a specific therapeutic domain.
    
    The framework then guides users through key reflection points:
    \begin{itemize}
        
    \item Is personalisation necessary and beneficial? And if so, what aspects do we need to optimise?
    
    Personalisation is used only to adapt exercise difficulty, reminder frequency, and encouragement strategies, avoiding unnecessary data collection about unrelated aspects and irrelevant adaptation. 
    
    \item Could the robot manipulate or foster overtrust?
    
    To mitigate these risks, the robot explains why recommendations are made and avoids emotional persuasion or excessive anthropomorphism.
    
    \item How is user autonomy preserved?
    
    Users can override recommendations, select alternative exercises, and temporarily disable adaptations. Therapists can review and adjust the robot's decisions.
    
    \item Could biased user models emerge?
    
    Performance estimates are continuously updated rather than assuming fixed cognitive abilities. Therapists periodically inspect the user model to detect systematic errors or outdated assumptions.
    
    \item How are privacy and data managed?
    
    Only task-relevant information is stored, users are informed about the collected data, and historical interaction records can be deleted upon request.
    
    \item Who is responsible for the system?
    
    Responsibilities are explicitly divided among developers, clinicians, and caregivers, and interaction logs are maintained to facilitate auditing and accountability.
    \end{itemize}
    
    This example shows how to transform abstract ethical principles into a practical decision-making process. Rather than asking how we can personalise as much as possible, the framework encourages the main stakeholders involved in the design process to ask whether personalisation is needed, what risks it introduces, and which safeguards should accompany it.

\section{Call to Actions}
Despite the implications discussed in the paper, current research in HRI largely evaluates personalisation in terms of efficiency, engagement, or short-term task success. The broader and longer-term consequences of adaptive systems remain comparatively underexplored. Rather than focusing exclusively on how to personalise a system, we argue that the field must return to a more fundamental question: when does personalisation meaningfully enhance human–robot interaction, and to what extent can it be implemented without generating unintended psychological, behavioural, or societal side effects?
We contend that the HRI community is at a critical juncture. The risks associated with personalisation should not be treated as peripheral ethical add-ons or as secondary consequences of technical implementation. Instead, they should be considered from the very beginning of the design process to properly address the double-edged nature of adaptive systems. To support this effort, we propose the establishment of an open community initiative dedicated to responsible personalisation in HRI.
This initiative will take the form of a publicly accessible platform \footnote{\url{https://sites.google.com/view/responsible-personalisation}} serving as a living repository of research questions, conceptual frameworks, current community initiatives, and shared resources. The platform will provide access to relevant publications, workshop materials, and recordings, while enabling researchers and practitioners to register their interest in contributing to an ongoing dialogue. Importantly, it will not advance a fixed normative position, but rather cultivate an evolving research agenda shaped through community participation.
In parallel, we will continue organising annual structured debates within our workshops, focusing on contested and emerging questions surrounding adaptive personalisation. These debates are intended to create space for rigorous, evidence-based discussion of issues such as autonomy preservation, persuasive adaptation, dependency risks, and transparency in embodied systems. Recognising the importance of democratising technological discourse, we also aim to produce accessible articles and public-facing materials to foster broader awareness of the adaptive technologies individuals increasingly integrate into their everyday lives.
We further invite the community to engage with the open research questions outlined in Section~\ref{sec:recommendations}, making them available for collaboration, discussion, and critical reflection. Addressing these questions will require sustained collaboration across HRI, psychology, ethics, law, and design research. By fostering an open and recurring forum for exchange, we aim to support the development of evaluation frameworks and design principles that balance adaptive performance with the preservation of human agency.
The future trajectory of personalisation in HRI should not be determined solely by technical feasibility or optimisation metrics. It should emerge from deliberate, evidence-driven, and community-wide reflection on what kinds of adaptive systems we seek to build and what forms of human–robot relationships we aim to cultivate.

\section*{Authors contributions}
A.A. and J.N. jointly led and coordinated the publication project, including conceptualisation of the manuscript, development of the overall structure, drafting and revising the manuscript, and preparation of the figures. A.R. contributed to the conceptualisation, structuring, and drafting of specific sections of the paper, and A.K. contributed to the drafting of specific sections of the paper. D.L., A.C., and T.L. contributed to parts of the manuscript. G.L. contributed to the overall structure of the paper and provided continuous feedback throughout the writing process. E.A. and V.C. reviewed drafts of the manuscript. All authors read and approved the final manuscript.

\section*{Acknowledgement}
In order to generate the Figures of this manuscript, AI tools were used. Specifically, chatgpt (GPT-5.5) together with draw.io was used for Figure~\ref{critical_questions}, figurelabs.ai was used to generate  Figure~\ref{Summative_figure}, whimsical.com was used to generate Figure~\ref{decision_framework}.

\bibliographystyle{plain}
\bibliography{./references.bib}

\end{document}